\documentclass[letterpaper]{article} 
\usepackage[preprint]{aaai2027}  
\usepackage[hyphens]{url}  
\usepackage{graphicx} 
\usepackage{natbib}  
\usepackage{caption} 
\usepackage{algorithm}

\usepackage{newfloat}
\usepackage{listings}

\usepackage{multirow}
\usepackage{amsmath,amsfonts,amssymb}
\usepackage{subcaption}
\usepackage{algpseudocode}
\usepackage{placeins}
\usepackage[capitalize,nameinlink,noabbrev]{cleveref}

\DeclareCaptionStyle{ruled}{labelfont=normalfont,labelsep=colon,strut=off} 
\floatstyle{ruled}
\newfloat{listing}{tb}{lst}{}
\floatname{listing}{Listing}

\crefname{figure}{Fig.}{Figs.}
\crefname{figure}{Fig.}{Figs.}
\crefname{section}{Sec.}{Secs.}
\crefname{section}{Section}{Sections}
\crefname{table}{Table}{Tables}
\crefname{table}{Tab.}{Tabs.}
\crefname{equation}{Eq.}{Eq.}

\usepackage{booktabs}

\title{Stop Replacing Noise with Noise: Two-Source Reliability Assessment for Label Correction and Sample Reweighting in Label-Noise Learning}
\author {
    Wenxiao Fan, Kan Li
}
\affiliations {
    School of Computer Science, Beijing Institute of Technology\\
    \{wenxiaofan, likan@bit.edu.cn\}
}

\begin{document}

\maketitle

\begin{abstract}
Refurbishment-based noisy-label learning mixes an observed label with a model-derived pseudo target, typically using one sample-wise cleanliness score to control both branches. This creates a hidden coupling: reducing trust in the observed label automatically increases trust in the pseudo target. We show that this complementarity can replace one unreliable signal with another because a pseudo target learned from corrupted supervision may reproduce the noise it is meant to correct. Our representation diagnostics provide a consistent account of this mismatch: noisy supervision redirects deeper layers more strongly, whereas shallower relations remain comparatively stable and provide information beyond the loss posterior. We therefore propose \textbf{TRACE}, a \textbf{T}wo-Source \textbf{R}eliability \textbf{A}ssessment framework for Label \textbf{C}orrection and Sample R\textbf{e}weighting. TRACE assesses the observed label using loss fit, shallow relation stability, and prediction agreement, while separately assessing the pseudo target using model confidence. Its source-specific scores control target correction and supervision strength without assuming complementary reliability. Across synthetic and real-world noisy benchmarks, TRACE improves representative refurbishment baselines and yields more reliable pseudo supervision.
\end{abstract}


\section{Introduction}

Learning with noisy labels remains a central challenge in modern supervised learning \citep{DBLP:conf/nips/NatarajanDRT13,DBLP:journals/corr/abs-2008-08186}. Deep networks can recover stable semantic structure from corrupted data, but they can also absorb erroneous supervision during optimization. The difficulty is not only to identify which labels are likely corrupted, but also to decide which supervision signal should guide each sample as training evolves.
Recent progress has followed several directions, including robust losses, sample selection, and label correction \citep{DBLP:conf/aaai/GhoshKS17,Han2018c,DBLP:conf/icml/ZhengWG0MC20}. Among them, refurbishment-based learning offers a practical interface for this decision: methods such as DivideMix \citep{DBLP:conf/iclr/LiSH20}, SELFIE \citep{Song2019}, SELC \citep{DBLP:conf/ijcai/LuH22}, and RoLR \citep{Chen2021TwoWD} construct a rectified or pseudo target and mix it with the observed noisy label during training.


Despite their differences, many refurbishment-based methods share the same control interface: they interpolate between the observed noisy label and a model-derived pseudo target using a single sample-wise cleanliness score \cite{Song2019,DBLP:conf/iclr/LiSH20,DBLP:conf/ijcai/LuH22,Chen2021TwoWD}, as formalized in \cref{eq:refurbishment_mix}. This interface collapses two questions into one scalar: how reliable is the observed label, and how reliable is the proposed pseudo target? Because the pseudo branch receives the complementary weight, distrusting the observed label automatically increases the influence of the pseudo target.

\begin{figure}[t]
  \centering
  \begin{subfigure}[t]{0.48\linewidth}
    \centering
    \includegraphics[width=1\linewidth]{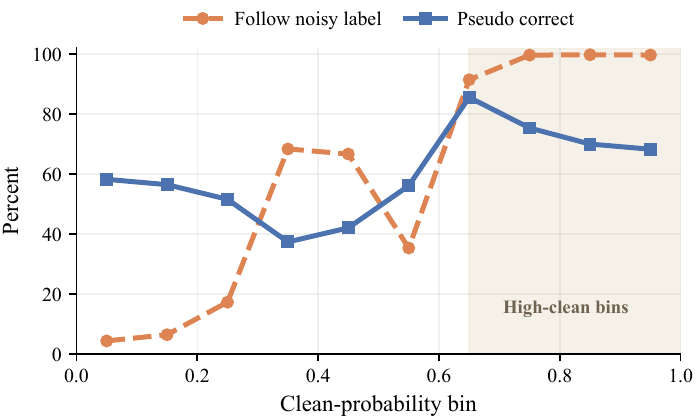}
    \caption{Bin-wise pseudo-target mismatch}
    \label{fig:intro_pseudo_target_unreliable_a}
  \end{subfigure}
  \hfill
  \begin{subfigure}[t]{0.48\linewidth}
    \centering
    \includegraphics[width=1\linewidth]{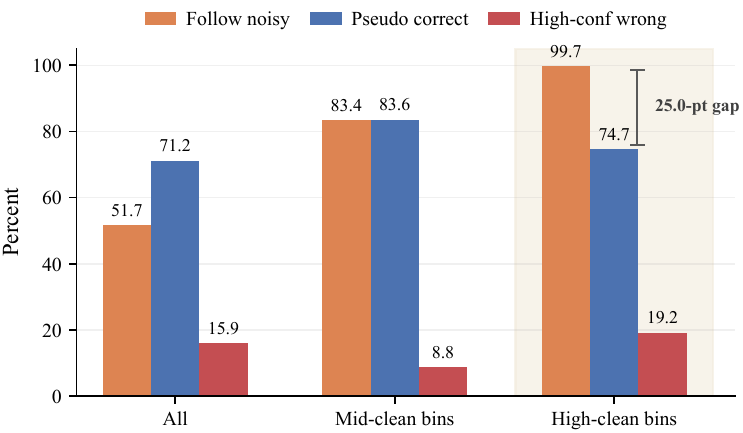}
    \caption{Aggregate failure across cleanliness regimes}
    \label{fig:intro_pseudo_target_unreliable_b}
  \end{subfigure}
  \caption{Pseudo-target unreliability under absorbed noisy bias on CIFAR-100N. Pseudo-target correctness is measured against the clean ground-truth label, and high sample cleanliness does not guarantee a trustworthy pseudo target.}
  \label{fig:intro_pseudo_target_unreliable}
   \vspace{-1.4em}
\end{figure}

\begin{figure*}[htbp]
  \centering
  \begin{minipage}[t]{0.675\textwidth}
    \centering
    \begin{minipage}[t]{0.32\linewidth}
      \centering
      \includegraphics[width=\linewidth]{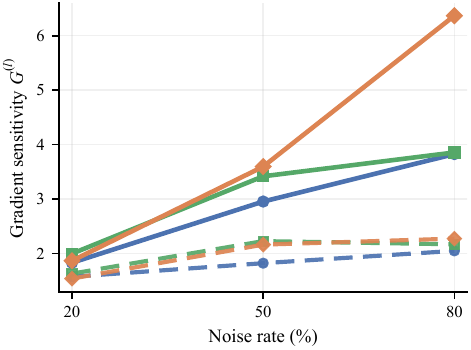}\\[-0.2em]
      {\scriptsize (i) Gradient sensitivity}
    \end{minipage}\hfill
    \begin{minipage}[t]{0.32\linewidth}
      \centering
      \includegraphics[width=\linewidth]{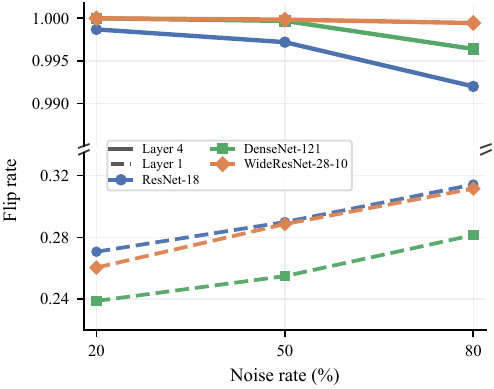}\\[-0.2em]
      {\scriptsize (ii) Prototype flipping}
    \end{minipage}\hfill
    \begin{minipage}[t]{0.32\linewidth}
      \centering
      \includegraphics[width=\linewidth]{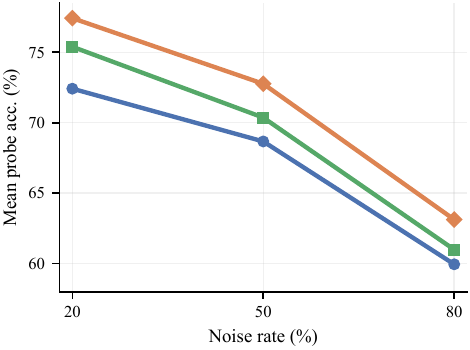}\\[-0.2em]
      {\scriptsize (iii) Linear-probe degradation}
    \end{minipage}
    \captionof{figure}{Depth-dependent noisy bias across ResNet-18, DenseNet-121, and WRN-28-10 on CIFAR-10 with 20\%, 50\%, and 80\% symmetric noise. Later layers show stronger gradient perturbation and prototype flipping, while linear-probe accuracy degrades as noise increases; the shared legend is shown in the middle plot.}
    \label{fig:preliminary_summary}
    \label{fig:preliminary_summary_a}
    \label{fig:preliminary_summary_b}
    \label{fig:preliminary_summary_c}
  \end{minipage}\hfill
  \begin{minipage}[t]{0.305\textwidth}
    \centering
    \includegraphics[width=\linewidth]{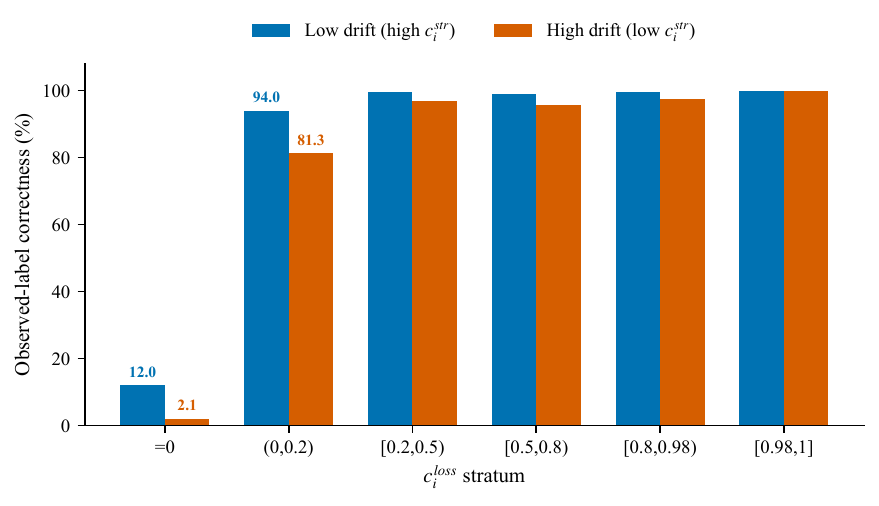}
    \captionsetup{justification=raggedright,singlelinecheck=false}
    \captionof{figure}{Relation drift complements loss confidence on CIFAR-100 with 50\% symmetric noise. Bars report observed-label correctness for low- and high-drift groups within each \(c_i^{\mathrm{loss}}\) stratum.}
    \label{fig:drift_comparison}
  \end{minipage}
 \vspace{-1.4em}
\end{figure*}

This coupling creates the risk of \emph{replacing noise with noise}. A pseudo target is not an external oracle; it is produced by a model trained on the same corrupted supervision. Once the model has absorbed noisy bias, the proposed correction can reproduce the error it is meant to replace, so observed-label and pseudo-target unreliability can coincide. \cref{fig:intro_pseudo_target_unreliable} gives a concrete example on CIFAR-100N \cite{DBLP:conf/iclr/WeiZ0L0022}: in high-clean bins, the pseudo target follows the noisy observed label 99.7\% of the time, yet is correct on only 74.7\% of samples and still makes 19.2\% high-confidence errors.
\cref{fig:clothing1m_pseudo_target_mismatch} shows the same qualitative mismatch on a randomly sampled Clothing1M subset. The two supervision sources must therefore be assessed with separate evidence.

Our representation analysis provides design evidence for this separation rather than a separate learning objective. Consistent with recent representation-level studies \cite{DBLP:conf/aistats/WongsoGM23,DBLP:conf/icml/MainiMSLKZ23}, noisy supervision perturbs deeper representations more strongly, pulls mislabeled samples toward noisy prototypes, and reduces the direct usability of deep features. This pattern helps explain why a pseudo target generated from a noise-trained deep state requires its own reliability assessment. Meanwhile, shallower relations remain comparatively stable, and \cref{fig:drift_comparison} shows that relation drift provides information about observed-label correctness beyond the loss posterior. Shallow structure can therefore refine observed-label reliability without certifying the pseudo target.

Motivated by this two-source view, we propose \textbf{TRACE}, a \textbf{T}wo-Source \textbf{R}eliability \textbf{A}ssessment framework for Label \textbf{C}orrection and Sample R\textbf{e}weighting. TRACE leaves the base pseudo-target generator unchanged but evaluates the observed label and pseudo target with separate reliability signals. The observed-label score combines loss fit, shallow relation stability, and prediction agreement; the pseudo-target score uses an independent confidence gate. Their composition enables need-aware pseudo supervision, while the resulting source-specific weights construct the corrected target and adjust each sample's contribution.


\noindent\textbf{Contributions.}
Our contributions are threefold: we identify the \emph{replacing-noise-with-noise} risk caused by using one cleanliness score to control both supervision sources; we propose TRACE, a plug-in two-source reliability framework for label correction and sample reweighting; and we show that TRACE improves refurbishment methods across synthetic, human-noise, and large-scale real-noise benchmarks.

\section{Related Work}

\textbf{Sample Selection and Label Refurbishment.}
Classical noisy-label learning often uses robust losses or sample selection, especially the small-loss effect, to separate clean from corrupted samples \citep{DBLP:conf/aaai/GhoshKS17,Han2018c,DBLP:conf/icml/ZhengWG0MC20}. Later methods enrich this decision with meta learning, dynamic correction, clean-noisy splitting, neighborhood/eigenvector cues, or confidence tracking \citep{DBLP:conf/cvpr/LiWZK19,DBLP:conf/icml/KimBZL23,DBLP:conf/cvpr/Li0S023,DBLP:journals/ijcv/KimRCK25,DBLP:journals/pr/CordeiroC25,DBLP:conf/aaai/Pan00D25}. A related line performs label correction/refurbishment by mixing observed labels with  pseudo targets \citep{DBLP:conf/iclr/LiSH20,Song2019,DBLP:conf/aaai/WuSX0M21,DBLP:conf/cvpr/TuZLLLWWZ23,DBLP:conf/ijcai/LuH22,Chen2021TwoWD}. TRACE is complementary: rather than another clean-sample detector or pseudo-target generator, it asks whether one scalar should control trust in both branches.

\noindent\textbf{Structure-Aware Reliability Estimation.}
Structure-aware methods use neighborhoods, contrastive relations, prototypes, and representation geometry to move beyond scalar losses \citep{Iscen2022LearningWN,Yi2022OnLC,DBLP:conf/cvpr/LiXGL22,DBLP:conf/cvpr/KarimRRMS22,DBLP:conf/nips/KimKCCY21,DBLP:conf/aistats/WongsoGM23,DBLP:conf/icml/MainiMSLKZ23}. They show that label noise appears in sample relations and layer-wise representations, not only in losses. TRACE uses shallow structural stability specifically to refine observed-label reliability, while a separate signal assesses the reliability of pseudo supervision from deeper states.

\begin{figure*}[t]
  \centering
  \includegraphics[width=0.93\linewidth]{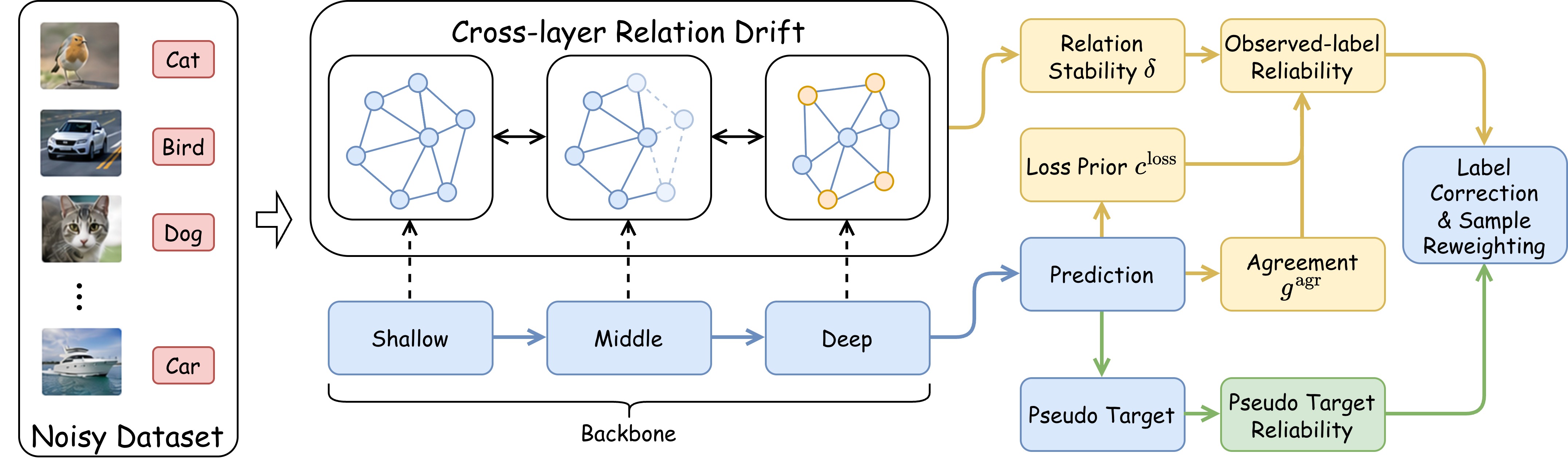}
  \caption{\textbf{Overview of TRACE.} TRACE assesses the observed label using loss confidence, cross-layer relation stability, and prediction agreement, while separately assessing the pseudo target using model confidence. The resulting source-specific scores drive label correction and sample reweighting.}
  \label{fig:framework}
  \vspace{-1.3em}
\end{figure*}

\noindent\textbf{Pseudo-Target Reliability Beyond Sample Cleanliness.}
Adjacent areas also recognize that pseudo labels or model-provided supervision can be unreliable and may require filtering or rectification \citep{Wang2024-nj,DBLP:journals/corr/abs-2403-06869,DBLP:journals/corr/abs-2410-07689}. These works support the broader intuition that pseudo supervision should not be treated as uniformly trustworthy. TRACE brings this distinction into standard refurbishment-based noisy-label classification: it separates observed-label reliability from pseudo-target reliability and ties the separation to depth-dependent noisy bias.

\section{Preliminary Analysis}
\label{sec:preliminary-analysis}

The two-source view requires separate evidence for observed-label and pseudo-target reliability. We use the following analyses for two supporting roles: the population-level shallow--deep contrast diagnoses why a noise-trained pseudo target cannot inherit reliability from the rejection of an observed label, while the sample-level relation-drift analysis tests whether shallow structure can refine observed-label reliability beyond the loss posterior. Clean labels and clean prototypes are used only for these diagnostics and are not required by TRACE during training.

\noindent\textbf{Deeper Layers Are More Easily Redirected by Noisy Supervision.}
We first quantify how label corruption perturbs training dynamics across depth. Let \(g_l^{\text{clean}}\) and \(g_l^{\text{noisy}}\) denote the average cross-entropy gradients of layer \(l\) under clean and noisy supervision on the same batch, and define
\begin{equation}
G^{(l)} = \frac{\left\|g_l^{\text{noisy}} - g_l^{\text{clean}}\right\|_2}{\left\|g_l^{\text{clean}}\right\|_2 + \varepsilon}
\end{equation}
Larger \(G^{(l)}\) indicates stronger noise-induced redirection. On ResNet-18 with CIFAR-10 symmetric noise, \(G^{(l)}\) rises from 1.5749 to 2.0510 at layer1 as the noise rate increases from 20\% to 80\%, but from 1.8334 to 3.8271 at layer4. \cref{fig:preliminary_summary_a} shows the same shallow--deep contrast for DenseNet-121 and WRN-28-10, indicating that deeper representations drift more strongly from comparatively stable shallow reference states.

\noindent\textbf{Mislabeled Samples Flip Toward Noisy Prototypes in Deep Representations.}
We next examine the geometric consequence through prototype attraction. For a mislabeled sample \(x\), let \(h_l(x)\) be its layer-\(l\) feature and \(\mu_l^{\text{clean}}(c)\) and \(\mu_l^{\text{obs}}(c)\) be class-\(c\) prototypes under clean and observed labels. We compare
\begin{equation}
  \begin{aligned}
    d_l^{\text{clean}}(x) &= \left\|h_l(x) - \mu_l^{\text{clean}}(y_{\text{clean}})\right\|_2\\
d_l^{\text{obs}}(x) &= \left\|h_l(x) - \mu_l^{\text{obs}}(y_{\text{obs}})\right\|_2\\
  \end{aligned}
\end{equation}
and summarize the tendency to align with the observed-label prototype by
\begin{equation}
\mathrm{flip\_rate}_l =
\mathbb{E}\big[\mathbf{1}\!\left[d_l^{\text{obs}}(x) < d_l^{\text{clean}}(x)\right]\big]
\end{equation}
On CIFAR-10 with 50\% symmetric noise, the flip rate rises from 29.0\% at layer1 to 99.7\% at layer4; on CIFAR-100, it rises from 28.3\% to 98.9\%. \cref{fig:preliminary_summary_b} shows the same late-stage flipping for DenseNet-121 and WRN-28-10. Thus, noise actively reorganizes deep features around incorrect class anchors, while the much lower shallow flip rate supports using shallow geometry to assess deep representation drift.

\begin{table*}[htbp]
  \centering
  \footnotesize
    \setlength{\tabcolsep}{3.5pt}
    \begin{tabular}{lccccccccccc}
    \toprule
    \textbf{Dataset} & \multicolumn{5}{c}{\textbf{CIFAR-10}} & \multicolumn{5}{c}{\textbf{CIFAR-100}} & \multirow{3}{*}{\textbf{Avg. \(\Delta\)}} \\
    \cmidrule(lr){2-6}\cmidrule(lr){7-11}
    \textbf{Noise Type} & \multicolumn{3}{c}{\textbf{Sym}}  & \textbf{Pair} &  \textbf{Ins}  & \multicolumn{3}{c}{\textbf{Sym}}  & \textbf{Pair}  & \textbf{Ins}   \\
    \cmidrule(lr){2-4}\cmidrule(lr){7-9}
    \textbf{Method / Noise Rate} & \textbf{20\%} & \textbf{50\%}& \textbf{80\%}& \textbf{40\%}  & \textbf{40\%}  & \textbf{20\%} & \textbf{50\%}& \textbf{80\%}& \textbf{40\%} & \textbf{40\%}    \\
    \midrule
     

    JoCoR \cite{DBLP:conf/cvpr/WeiFC020}  & 89.4  & 53.3  & 25.8  &56.1 & 60.9  & 55.4  & 32.7  & 6.6 & 34.1 & 34.9 & --\\

    RankMatch \cite{DBLP:conf/iccv/Zhang0FLCLL23} & 96.4 &	95.4&	94.2&  94.4 & 93.8  & 79.3&	77.6&	67.2&  75.8& 76.5 & -- \\
    CrossSplit \cite{DBLP:conf/icml/KimBZL23} & 96.9	&96.3	&95.4			&96.0 & 95.8 &79.9& 75.7 & 64.6&76.8 & 79.2 & -- \\
    CCL \cite{DBLP:conf/aaai/FanL25} &  {97.0}&	{\underline{96.5}}&	{94.6}&  {96.1}	&	{96.2} & {79.5}	&{77.4}&	{70.3} & {77.2}	 &	{\underline{80.0}} & -- \\
    NegScale \cite{DBLP:conf/aaai/FanL26} & {\underline{97.2}} & {\textbf{96.6}} & {\underline{95.6}} & {\underline{96.3}} & \underline{96.5} & {\underline{80.9}} & {\underline{78.7}} & {70.8} & {\underline{77.8}} & \textbf{80.4} & -- \\
    \midrule

    DivideMix \cite{DBLP:conf/iclr/LiSH20}& 95.7 &	94.4&	92.9	&	92.1	&	95.1& 76.9	&74.2&	59.6&	52.3&	76.1 & --\\
    \quad + TRACE & 96.1 & 94.9 & 93.5 & 92.6 & 95.5 & 78.0 & 75.1 & 61.0 & 53.8 & 77.0 & \textbf{+0.82} \\
    RoLR \cite{Chen2021TwoWD} & 96.4&	95.7&	94.2&	92.8 & 	93.7&  78.6&	74.6&	66.2& 76.1 & 77.2 & -- \\
     \quad + TRACE &97.0 & 96.2 & 94.7 & 93.9 & 94.1 & 80.4 & 76.2 & 68.0 & \textbf{78.1} & 78.5 & \textbf{+1.16} \\
    DISC \cite{DBLP:conf/cvpr/Li0S023} & 96.3 &95.4	&92.9&94.6	&	96.0 & 78.6	&76.3&	59.3&75.1  &78.4 & -- \\
     \quad + TRACE & 96.7 & 95.8 & 93.4 & 95.0 & 96.3 & 79.4 & 77.0 & 60.4 & 76.0 & 79.1 & \textbf{+0.62} \\
    ANNE \cite{DBLP:journals/pr/CordeiroC25} &96.9 & 96.2 & 95.3 & 95.7 &96.2 & 80.4 & 78.1 & {\underline{73.0}} & 66.4 & 78.4 & -- \\
     \quad + TRACE & \textbf{97.4} & \textbf{96.6} & \textbf{95.9} & \textbf{96.4} & \textbf{96.6} & \textbf{81.2} & \textbf{78.8} & \textbf{74.2} & 67.2 & 79.3 & \textbf{+0.70} \\
    \bottomrule
    \end{tabular}%
      \caption{Comparison with state-of-the-art methods on CIFAR-10/100 datasets under various types of noise. 
    The results of other methods are from the published results of corresponding papers.
    Bold and underline denote the best and second-best results, respectively.
    Avg. \(\Delta\) reports the mean improvement of each TRACE plug-in over its paired base learner across the ten settings.
     }
     \vspace{-1em}
  \label{tab:main_1}%
\end{table*}%

\noindent\textbf{Prototype Flipping Reduces the Direct Usability of Deep Representations.}
Finally, we test whether prototype flipping reduces the direct usability of learned features. We freeze \(h_l(x)\) at each depth, fit a linear classifier \(W_l\), and measure
\begin{equation}
\mathrm{ProbeAcc}_l =
\frac{1}{|\mathcal{D}_{\text{test}}|}
\sum_{(x,y)\in\mathcal{D}_{\text{test}}}
\mathbf{1}\!\left[\arg\max W_l h_l(x) = y\right]
\end{equation}
The layer1--4 mean probe accuracy drops from 72.42\% to 59.95\% on CIFAR-10 and from 42.79\% to 28.48\% on CIFAR-100 as symmetric noise increases from 20\% to 80\%. \cref{fig:preliminary_summary_c} shows the same monotonic decline for DenseNet-121 and WRN-28-10. Hence, rejecting an observed label does not justify trusting a pseudo target produced from deep representations that may no longer align with clean semantics.

\noindent\textbf{Relation Drift Provides Information Beyond Loss Confidence.}
The preceding analyses establish a population-level shallow--deep contrast; TRACE, however, needs a sample-level signal. We therefore compare low- and high-drift samples within the same \(c_i^{\mathrm{loss}}\) strata on CIFAR-100 with 50\% symmetric noise. \cref{fig:drift_comparison} shows that low-drift samples more often retain the clean label, especially when loss-based evidence is weak, with the gap narrowing as \(c_i^{\mathrm{loss}}\) approaches one. Relation drift therefore provides conditional information beyond the loss posterior and is used as a complementary modifier rather than a standalone correctness certificate.

Together, \cref{fig:intro_pseudo_target_unreliable,fig:preliminary_summary,fig:drift_comparison} motivate two distinct reliability assessments. Observed-label unreliability does not establish pseudo-target reliability, and the depth-dependent diagnostics provide a consistent account of why the pseudo branch needs independent evidence. Relation drift, in contrast, supplies information beyond the loss posterior for refining observed-label reliability. TRACE therefore uses shallow-anchored stability only in the observed-label score and filters pseudo supervision through a separate reliability signal.

\section{Method}
\label{sec:method}

TRACE reformulates refurbishment as separate reliability assessments for its two supervision sources: the observed label and the model-derived pseudo target. As shown in \cref{fig:framework}, the observed-label score combines the loss prior, shallow-to-deep relation stability, and dual-network agreement, while the pseudo-target score uses an independent confidence signal. Their composition preserves the base pseudo-target generator but replaces automatic branch complementarity with need-aware, independently gated pseudo supervision. The complete mini-batch procedure is given in \cref{alg:trace_pseudocode}.

\noindent\textbf{Setup and Two-Source Reliability.}
Let \(\mathcal{D} = \{(x_i, \hat{\mathbf{y}}_i)\}_{i=1}^{N}\), where \(\hat{\mathbf{y}}_i \in \{0,1\}^{C}\) is the one-hot observed label, and let \(p_i = f_{\theta}(x_i) \in \Delta^{C-1}\) be the model prediction. A refurbishment learner also constructs a pseudo target \(q_i \in \Delta^{C-1}\); in our RoLR-style instantiation,
\begin{equation}
q_i = \text{Sharpen}_{T}\!\left(\frac{p_i^{(1)} + p_i^{(2)}}{2}\right)
\end{equation}
where \(p_i^{(1)}\) and \(p_i^{(2)}\) are predictions from two networks. Standard methods mix the observed and pseudo-target branches through
\begin{equation}
\label{eq:refurbishment_mix}
\mathbf{y}_i^{\text{rec}} = \lambda_i \hat{\mathbf{y}}_i + (1-\lambda_i) q_i,
\qquad \lambda_i \in [0,1]
\end{equation}
Following DivideMix and RoLR~\citep{DBLP:conf/iclr/LiSH20,Chen2021TwoWD}, the observed-label coefficient \(\lambda_i\) is commonly derived from the loss-based clean posterior
\begin{equation}
c_i^{\text{loss}} = P(z_i=\text{clean} \mid \bar{\ell}_i)
\end{equation}
where \(z_i \in \{\text{clean},\text{noisy}\}\) is the latent clean state and \(\bar{\ell}_i\) is the normalized cross-entropy loss. Because \(1-\lambda_i\) simultaneously weights the pseudo branch, the same signal couples the two reliability decisions and treats reduced observed-label trust as increased pseudo-target trust.

TRACE instead introduces \(s_i^{\text{obs}}\) for observed-label reliability and \(s_i^{\text{pseudo}}\) for pseudo-target reliability, and composes them into the branch weights
\begin{equation}
\label{eq:weights}
a_i = s_i^{\text{obs}},
\qquad
b_i = \big(1 - s_i^{\text{obs}}\big) s_i^{\text{pseudo}}.
\end{equation}
The observed branch is weighted directly by its reliability score. The pseudo branch is decomposed into two factors: \(1-s_i^{\text{obs}}\) expresses the need for correction, whereas \(s_i^{\text{pseudo}}\) measures the reliability of the proposed correction. The two branch weights are therefore coordinated but not complementary: low observed-label reliability can create a need for pseudo supervision, but it cannot activate that supervision without independent pseudo-target evidence.

\begin{table*}[htbp]
  \centering
    \footnotesize
    \setlength{\tabcolsep}{4pt}
    \begin{tabular}{llccccccc}
    \toprule
    \multicolumn{2}{l}{\textbf{Dataset}} & \multicolumn{5}{c}{\textbf{CIFAR-10N}} & \textbf{CIFAR-100N} & \multirow{3}{*}{\textbf{Avg. \(\Delta\)}} \\
    \cmidrule(lr){3-7}\cmidrule(lr){8-8}
    \multirow{2}{*}{\textbf{Method}}  & \textbf{Noise Type} & \textbf{Aggre} & \textbf{Rand1} & \textbf{Rand2} & \textbf{Rand3} & \textbf{Worst} & \textbf{Fine}   \\
          & \textbf{Noise Rate} & \textbf{9.0}\% & \textbf{17.2\%} &\textbf{18.12\%} &\textbf{17.64}\%& \textbf{40.2\%}& \textbf{40.2\%} \\
        \midrule
    
    \multicolumn{2}{l}{Cross-Entropy} & 88.8  & 83.8  & 83.5  & 83.9  & 67.7  & 47.8   \\

    \multicolumn{2}{l}{Co-teaching  \cite{Han2018c}} & 89.9  & 87.8  & 87.2  & 87.4  & 62.3  & 40.5 & --  \\
    {JoCoR  \cite{DBLP:conf/cvpr/WeiFC020}} & & 90.6  & 88.8  & 88.5  & 88.1  & 66.7  & 40.1 & --  \\
        \multicolumn{2}{l}{Co-learning   \cite{DBLP:conf/mm/TanXWL21}} & 92.4  & 91.3  & 91.2  & 91.4  & 81.0  & 47.9   \\
     RankMatch \cite{DBLP:conf/iccv/Zhang0FLCLL23}& & 95.6 & 94.8 & 95.1 & 95.3 & 92.8 & 65.2 & -- \\
    CCL \cite{DBLP:conf/aaai/FanL25} & & {96.4}&	{\underline{96.0}}&	{95.8}	&{96.1}	&{\underline{93.1}}	& {65.5} & -- \\
    NegScale \cite{DBLP:conf/aaai/FanL26} & & \underline{96.6} & \textbf{96.2} & \underline{96.0} & \underline{96.4} & \textbf{93.5} & \underline{66.3} & -- \\
    \midrule
    \multicolumn{2}{l}{DivideMix \cite{DBLP:conf/iclr/LiSH20}}&93.2 & 92.8 &92.6 & 93.1 & 89.2 & 55.2 & -- \\
    \quad + TRACE && 93.6 & 93.2 & 93.1 & 93.5 & 89.8 & 56.0 & \textbf{+0.52} \\
     RoLR \cite{Chen2021TwoWD} & & 95.4 & 94.9 & 94.7 & 95.2 & 92.3 & 62.3 & -- \\
     \quad + TRACE && 95.8 & 95.3 & 95.2 & 95.6 & 92.8 & 63.1 & \textbf{+0.50} \\
    ANNE \cite{DBLP:journals/pr/CordeiroC25} & & 96.2 & 95.7 & 95.5 & 95.9 & 93.0 & 66.0 & -- \\
    \quad + TRACE && \textbf{96.8} & \textbf{96.2} & \textbf{96.1} & \textbf{96.6} & \textbf{93.5} & \textbf{66.6} & \textbf{+0.58}\\


    \bottomrule
    \end{tabular}%
        \caption{Comparison with state-of-the-art methods on CIFAR-N. The results are from \cite{Wei2022MitigatingMO} and our replication. Bold and underline denote the best and second-best results, respectively. 
    }
  \label{tab:main_3}%
  \vspace{-0.8em}
\end{table*}%

\noindent\textbf{Observed-Label Reliability.}
We first estimate \(s_i^{\mathrm{obs}}\), the reliability of the observed label. \cref{fig:drift_comparison} shows that relation drift separates samples with different observed-label correctness within the same loss-confidence strata. We therefore combine \(c_i^{\mathrm{loss}}\), which captures the small-loss prior, with \(c_i^{\mathrm{str}}\), which measures shallow-to-deep relational stability. In particular, hard-but-clean samples near class boundaries may also undergo natural cross-layer drift, as shown in Appendix.B. Therefore, \(c_i^{\mathrm{str}}\) only refines the loss posterior; it is neither a standalone correctness estimator nor a direct certificate for the pseudo target.

For mini-batch features \(\{h_l(x_i)\}_{i=1}^{B}\) at layer \(l\), we compute the cosine relation matrix
\begin{equation}
A_{ij}^{(l)} =
\left\langle
\frac{h_l(x_i)}{\|h_l(x_i)\|_2},
\frac{h_l(x_j)}{\|h_l(x_j)\|_2}
\right\rangle
\end{equation}
To retain dominant local relations, let \(\mathcal{N}_k(i)\) contain sample \(i\) and its \(k\) largest off-diagonal affinities, and set \(M_{ij}^{(l)}=\mathbf{1}[j\in\mathcal{N}_k(i)]\). The symmetric local relation matrix \(R^{(l)}=\mathcal{S}_k(A^{(l)})\) is
\[
\mathcal{S}_k(A^{(l)}) =
\frac{1}{2}\left(A^{(l)}\odot M^{(l)}
+ \left(A^{(l)}\odot M^{(l)}\right)^\top\right)
\]
The row \(R_i^{(l)}\) describes sample \(i\)'s local position at depth \(l\). Its cross-layer change defines relation drift and structure confidence:
\begin{equation}
  \begin{aligned}
    \delta_i &= \sum_{l=1}^{L-1} \frac{\left\|R_i^{(l+1)} - R_i^{(l)}\right\|_2}{\sqrt{B}}\\
\tilde{\delta}_i &= \operatorname{Norm}(\delta_i)\\
c_i^{\text{str}} &= \operatorname{Norm}\!\big(\exp(-\gamma\tilde{\delta}_i)\big)
  \end{aligned}
\label{eq:structure_confidence}
\end{equation}
Here \(L\) is the number of analyzed layers, \(B\) is the batch size, \(\gamma>0\) maps drift to confidence, and \(\operatorname{Norm}(\cdot)\) is min-max normalization over the current scoring pass. A larger \(c_i^{\text{str}}\) indicates greater relational stability. For two networks, we use \(\bar{c}_i^{\text{str}}=\tfrac{1}{2}(c_{i,1}^{\text{str}}+c_{i,2}^{\text{str}})\).

As depicted in \cref{fig:framework}, prediction agreement provides a lightweight consistency gate:
\begin{equation}
g_i^{\text{agr}} =
\begin{cases}
1, & \arg\max p_i^{(1)} = \arg\max p_i^{(2)},\\
\lambda_{\mathrm{dis}}, & \text{otherwise}
\end{cases}
\end{equation}
where \(\lambda_{\mathrm{dis}} \in (0,1]\) penalizes disagreement. The observed-label score combines this gate with the loss prior and \cref{eq:structure_confidence}:
\begin{equation}
  \begin{aligned}
    \alpha_t &= 1 - \beta_t(1-\alpha) \\
    s_i^{\text{obs}} &=
\operatorname{clip}\!\left(
\big(\alpha_t c_i^{\text{loss}} + (1-\alpha_t)\bar{c}_i^{\text{str}}\big)
\big((1-\beta_t) + \beta_t g_i^{\text{agr}}\big), 0, 1\right)
  \end{aligned}
\end{equation}
where \(\alpha \in [0,1]\) sets the final loss--structure balance and \(\beta_t\) gradually activates structure and agreement. Thus, observed-label reliability is assessed jointly by label fit, representation stability, and prediction agreement.

\noindent\textbf{Pseudo-Target Reliability.}
We next estimate \(s_i^{\mathrm{pseudo}}\), the reliability of \(q_i\) as an alternative supervision signal. This assessment requires its own evidence because low \(s_i^{\mathrm{obs}}\) indicates a need for correction but does not establish that the available pseudo target is reliable (\cref{fig:intro_pseudo_target_unreliable}). We score whether \(q_i\) is sufficiently certain to provide alternative supervision by
\begin{equation}
s_i^{\text{pseudo}} = \left(\max_{c} q_{i,c}\right)^{\rho}
\end{equation}
where \(\rho \ge 1\) controls the suppression of low-confidence targets. The gate is activated after a short warm-up; it filters uncertain replacements but is not treated as a correctness certificate.

\noindent\textbf{Label Correction and Sample Reweighting.}
Using the source-specific weights in \cref{eq:weights}, the rightmost block in \cref{fig:framework} constructs the corrected target
\begin{equation}
\tilde{\mathbf{y}}_i =
\frac{a_i \hat{\mathbf{y}}_i + b_i q_i}{a_i + b_i + \varepsilon}
\end{equation}
where \(\varepsilon > 0\) ensures numerical stability. The supervision strength and training loss are
\begin{equation}
\label{eq:supervision_weighted_loss}
\mathcal{L} = \frac{1}{N}\sum_{i=1}^{N}
w_i\,\ell\!\left(p_i,\tilde{\mathbf{y}}_i\right),w_i = \max(a_i + b_i, w_{\min})
\end{equation}
where \(w_{\min} \ge 0\) is a lower bound and \(\ell(\cdot,\cdot)\) is soft-target cross-entropy. In each mini-batch, \(w_i\) is normalized by its mean and the base prior regularizer is retained.
Consequently, reliable observed labels remain dominant, pseudo supervision increases only when correction is needed and the pseudo target is independently assessed as reliable, and samples receive less weight when neither source is reliable.

\noindent\textbf{Decision Roles and Scope.}
TRACE decouples the evidence used to evaluate the two supervision sources, not the training roles of their final weights. The pseudo branch remains need-aware through \(1-s_i^{\mathrm{obs}}\), but its influence is independently controlled by \(s_i^{\mathrm{pseudo}}\). The resulting controller separates observed-label reliability, pseudo-target reliability, and sample weighting without treating either score as a correctness certificate. \cref{fig:drift_comparison} and Appendix~\ref{app:reliability_view} provide the corresponding empirical and analytical motivation.

\begin{figure*}[htbp]
  \centering
  \begin{subfigure}[t]{0.245\textwidth}
    \centering
    \includegraphics[width=\linewidth]{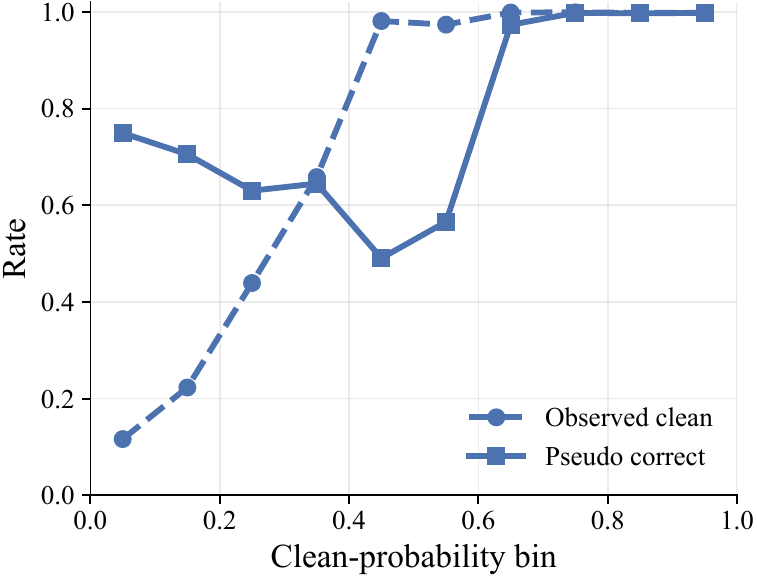}
    \caption{Reliability by bin}
    \label{fig:exp_pseudo_reliability_a}
  \end{subfigure}\hfill
  \begin{subfigure}[t]{0.245\textwidth}
    \centering
    \includegraphics[width=\linewidth]{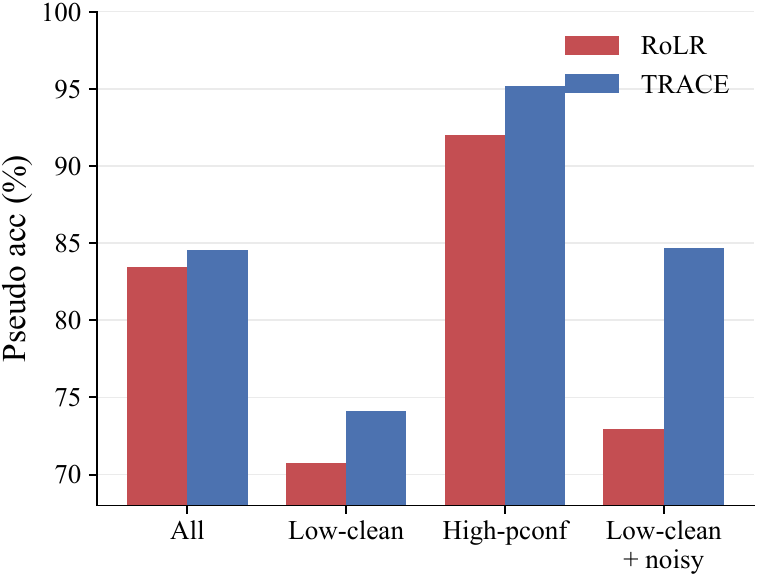}
    \caption{Pseudo-target acc.}
    \label{fig:exp_pseudo_reliability_b}
  \end{subfigure}\hfill
  \begin{subfigure}[t]{0.245\textwidth}
    \centering
    \includegraphics[width=\linewidth]{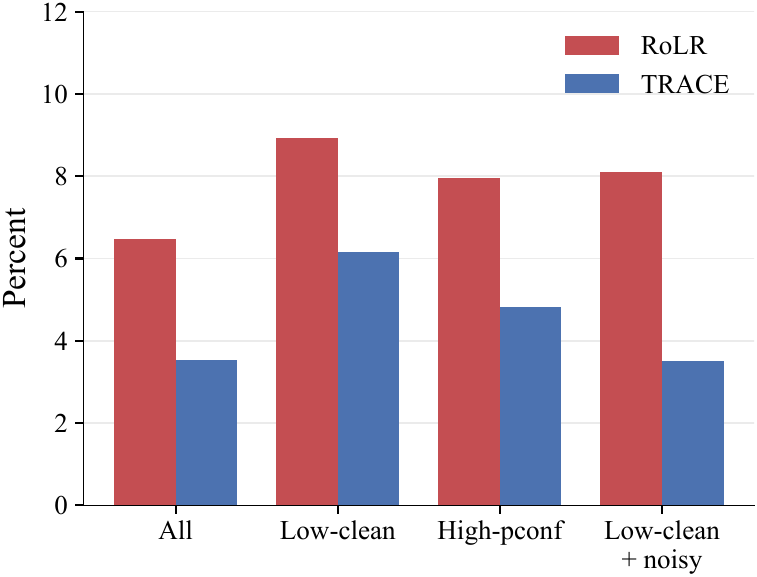}
    \caption{High-conf. errors}
    \label{fig:exp_pseudo_reliability_c}
  \end{subfigure}\hfill
  \begin{subfigure}[t]{0.245\textwidth}
    \centering
    \includegraphics[width=\linewidth]{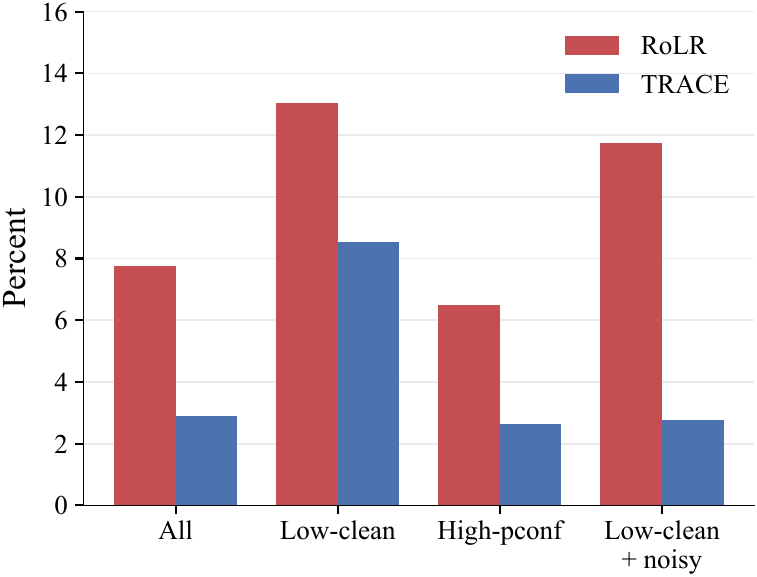}
    \caption{Calibration error}
    \label{fig:exp_pseudo_reliability_d}
  \end{subfigure}
  \caption{Effectiveness of TRACE on CIFAR-100 with 50\% symmetric noise: improved pseudo-target reliability with fewer high-confidence errors and lower calibration error.}
  \label{fig:exp_pseudo_reliability}
  \vspace{-0.6em}
\end{figure*}

\begin{table}[t]
  \centering
  \footnotesize
  \setlength{\tabcolsep}{1.6pt}
  \begin{tabular}{@{}lcccccc@{}}
    \toprule
    \multirow{2}{*}{\textbf{Method}} & \multicolumn{2}{c}{\textbf{WebVision}} & \multicolumn{2}{c}{\textbf{ILSVRC12}} & \textbf{Food-101N} & \multirow{2}{*}{\textbf{Avg. \(\Delta\)}} \\
    \cmidrule(lr){2-3}\cmidrule(lr){4-5}
     & \textbf{Top1} & \textbf{Top5} & \textbf{Top1} & \textbf{Top5} & \textbf{Acc.} & \\
    \midrule
    CORES & 71.7 & 89.0 & 68.4 & 88.3 & 84.4 & -- \\
    \quad + CT & 72.1 & 90.2 & 68.5 & 90.0 & 84.4 & +0.68 \\
    \quad + TRACE& 72.3 & 90.2 & 68.6 & 90.1 & 84.7 & \textbf{+0.82} \\
    \midrule
    DivideMix & 77.4 & 91.9 & 74.7 & 92.1 & 86.5 & -- \\
    \quad + CT & 78.1 & 92.3 & 75.2 & 92.2 & 86.8 & +0.40 \\
    \quad + TRACE& 78.2 & 92.3 & 75.3 & 92.3 & 86.8 & \textbf{+0.46} \\
    \midrule
    f-DivideMix & 78.4 & 92.5 & 75.2 & 92.2 & 86.8 & -- \\
    \quad + CT & 78.8 & 92.9 & 75.7 & 93.2 & 87.0 & +0.50 \\
    \quad + TRACE& 78.9 & \textbf{93.2} & 75.8 & \textbf{93.3} & 87.1 & \textbf{+0.64} \\
    \midrule
    DISC & 80.1 & 92.4 & 77.4 & 92.4 & 87.3 & -- \\
    \quad+ CT & 80.1 & 92.6 & \textbf{78.3} & 92.4 & 87.5 & +0.26 \\
    \quad + TRACE &\textbf{80.4} & 92.6 & \textbf{78.3} & 92.5 & \textbf{87.8} & \textbf{+0.48} \\
    \bottomrule
  \end{tabular}
    \caption{Average test accuracy (\%) over the last 10 epochs on large-scale real-world noisy-label datasets. 
  }
  \label{tab:large_scale_real_noise}
  \vspace{-1.4em}
\end{table}

\begin{table}[t]
  \centering
  \footnotesize
  \setlength{\tabcolsep}{0.8pt}
  \begin{tabular}{@{}lccccccc@{}}
    \toprule
    \multirow{2}{*}{\textbf{Plug-in}} & \multicolumn{3}{c}{\textbf{CIFAR-10-Sym}} & \multicolumn{3}{c}{\textbf{CIFAR-100-Sym}} & \multirow{2}{*}{\textbf{Clothing1M}} \\
    \cmidrule(lr){2-4}\cmidrule(lr){5-7}
     & \textbf{20\%} & \textbf{50\%} & \textbf{80\%} & \textbf{20\%} & \textbf{50\%} & \textbf{80\%} & \\
    \midrule
    DivideMix & 95.7 & 94.4 & 92.9 & 76.9 & 74.2 & 59.6 & 74.8 \\
    \midrule
    \quad+ SNSCL & \textbf{96.1} & \textbf{95.2} & 91.7 & 77.3 & 74.7 & \textbf{64.3} & 75.3 \\
    \quad+ L2B & 95.9 & 95.1 & 93.2 & 77.3 & 74.9 & 60.2 & 76.1 \\
    \midrule
    \quad+ TRACE & \textbf{96.1} & 94.9 & \textbf{93.5} & \textbf{78.0} & \textbf{75.1} & 61.0 & \textbf{77.6} \\
    \bottomrule
  \end{tabular}%
    \caption{Test accuracy (\%) comparison with plug-in noisy-label methods.}
  \label{tab:plugin_method_comparison}
  \vspace{-1.4em}
\end{table}

\section{Experiments}

We evaluate TRACE from three complementary views: overall accuracy and plug-in generality, the contribution of two-source reliability assessment, and the quality of pseudo supervision.

\noindent\textbf{Experimental Setup.}
We evaluate TRACE on synthetic CIFAR noise and real-world CIFAR-10N/CIFAR-100N, WebVision, Food-101N, and Clothing1M \citep{krizhevsky2009learning,DBLP:conf/iclr/WeiZ0L0022,DBLP:journals/corr/abs-1708-02862,lee2018cleannet}. CIFAR experiments use ResNet-18, and the large-scale real-noise experiments use ResNet-50. For each run, test accuracy is averaged over the final 10 epochs, and the reported results are further averaged across three random seeds (0, 42, and 1027). TRACE preserves each base learner's pseudo-target generator, training schedule, and data pipeline, so each paired comparison changes only the reliability mechanism. Appendix~A provides the full protocols and hyperparameters.


\noindent\textbf{CIFAR synthetic noise.}
\cref{tab:main_1} covers symmetric, pair/asymmetric, and instance-dependent noise on CIFAR-10/100. TRACE improves every paired base learner in all reported settings, with generally larger gains on CIFAR-100. This pattern is consistent with source-specific reliability assessment becoming more useful when the larger label space makes pseudo supervision harder. ANNE+TRACE gives the strongest overall results, while RoLR+TRACE improves every RoLR result.

\noindent\textbf{Real-world noisy datasets.}
\cref{tab:main_3,tab:large_scale_real_noise} show that TRACE improves all paired results on CIFAR-N and improves or matches every reported WebVision and Food-101N metric. The gains span four base learners with different pseudo-target generators, rather than depending on one refurbishment rule. On Clothing1M (\cref{tab:plugin_method_comparison}), DivideMix+TRACE reaches 77.6\%, compared with 74.8\% for DivideMix.

\begin{table}
   \centering
    \footnotesize
    \setlength{\tabcolsep}{2.0pt}
    \begin{tabular}{lcc}
      \toprule
      \textbf{Variant} & \textbf{Best} & \textbf{Last} \\
      \midrule
      RoLR (base) & 75.68 & 74.71 \\
      Refined \(s_i^{\mathrm{obs}}\) only & 76.22 & 75.43 \\
      Decoupled w/o \(c_i^{\mathrm{str}}\) &76.51  & 75.78 \\
      Decoupled w/o \(g_i^{\mathrm{agr}}\) &76.43 &  75.70 \\
      \shortstack[l]{Coupled null 
      (\(b_i=1-s_i^{\mathrm{obs}}\))} & 76.58 & 75.86 \\
      \shortstack[l]{Decoupled reliability 
      (\(b_i=(1-s_i^{\mathrm{obs}})s_i^{\mathrm{pseudo}}\))} & {76.89} & {76.10} \\
      \midrule
      TRACE  & \textbf{76.95} & \textbf{76.22} \\
      \bottomrule
    \end{tabular}
        \caption{Method-component ablation in test accuracy (\%) on CIFAR-100 with 50\% symmetric noise.}
    \label{tab:main_ablation}
    \vspace{-0.6em}
\end{table}

\begin{table}
   \centering
    \footnotesize
    \begin{tabular}{lccc}
      \toprule
      \textbf{Setting} & \textbf{RoLR} & \textbf{TRACE} & \textbf{Gain} \\
      \midrule
      CIFAR100-Sym20 & 76.51 & 93.02 & +16.51 \\
	      CIFAR100-Sym50 & 72.92 & 84.70 & +11.78 \\
	      CIFAR100-Sym80 & 42.41 & 54.09 & +11.68 \\
      CIFAR-100N & 59.68 & 60.38 & +0.70 \\
	      CIFAR10-Sym50 & 96.42 & 97.52 & +1.10 \\
	      CIFAR-10N-Agg & 93.02 & 94.07 & +1.05 \\
	      CIFAR-10N-Worst & 93.83 & 95.78 & +1.95 \\
	      \bottomrule
	    \end{tabular}
      \caption{Pseudo-target accuracy (\%) on low-clean+noisy samples. 
    }
    \label{tab:pseudo_reliability_summary}
    \vspace{-1.4em}
\end{table}

\noindent\textbf{Comparison with plug-in methods.}
\cref{tab:large_scale_real_noise} compares TRACE with CT under matched base-learner protocols, and \cref{tab:plugin_method_comparison} compares DivideMix-based plug-ins.
Because the latter methods use different evaluation protocols, we report their displayed values as contextual comparisons rather than treating them as a strictly controlled leaderboard. The unmodified DivideMix row is the common reference, whereas the CT comparison keeps the base learner and evaluation protocol matched.
TRACE exceeds CT's average gain for every matched base learner, improves DivideMix in all CIFAR columns, and gives the highest reported Clothing1M accuracy.

\begin{table}[htbp]
  \footnotesize
      \centering
  \setlength{\tabcolsep}{4pt}
  \begin{tabular}{lccccc}
    \toprule
    \textbf{Factor} & \textbf{Value} & \textbf{Best} & \textbf{Last} & \textbf{Pseudo Acc.} & \textbf{HC Wrong} \\
    \midrule
     & \(0.5\) & 76.72 & 75.98 & 84.31 & 6.42 \\
    \(\alpha\) & \(0.7\) & \textbf{76.95} & \textbf{76.22} & \textbf{84.70} & \textbf{6.16} \\
     & \(0.9\) & \underline{76.83} & \underline{76.07} & \underline{84.48} & \underline{6.28} \\
     \midrule
     & \(0.5\) & 76.78 & 76.01 & 84.18 & 6.55 \\
    \(\rho\) & \(1.0\) & \textbf{76.95} & \textbf{76.22} & \textbf{84.70} & \underline{6.16} \\
     & \(1.5\) & \underline{76.87} & \underline{76.14} & \underline{84.56} & \textbf{6.05} \\
    \bottomrule
  \end{tabular}
  \caption{Main sensitivity analysis with loss--structure balance \(\alpha\) and pseudo-confidence power \(\rho\) on CIFAR-100 with 50\% symmetric noise. HC Wrong denotes the percentage of high-confidence pseudo targets that are incorrect (lower is better).}
  \label{tab:sensitivity_hyperparameters}
    \vspace{-0.6em}
\end{table}

\noindent\textbf{Ablation on two-source reliability.}
\cref{tab:main_ablation} isolates \(s_i^{\mathrm{obs}}\), its \(c_i^{\mathrm{str}}\) and \(g_i^{\mathrm{agr}}\) terms, \(s_i^{\mathrm{pseudo}}\), and the final loss on CIFAR-100 with 50\% symmetric noise.

Replacing RoLR's cleanliness score with \(s_i^{\mathrm{obs}}\) raises best/last accuracy from 75.68\%/74.71\% to 76.22\%/75.43\%. The coupled-null and decoupled-reliability variants then hold \(s_i^{\mathrm{obs}}\), pseudo targets, and cross-entropy fixed, changing only \(b_i\) from \(1-s_i^{\mathrm{obs}}\) to \((1-s_i^{\mathrm{obs}})s_i^{\mathrm{pseudo}}\). The gain from 76.58\%/75.86\% to 76.89\%/76.10\% therefore isolates the pseudo-target gate. Removing \(c_i^{\mathrm{str}}\) or \(g_i^{\mathrm{agr}}\) tests the two observed-label components, and restoring the supervision weight in \cref{eq:supervision_weighted_loss} yields the full 76.95\%/76.22\% result.
The progression shows that refining observed-label reliability and independently screening pseudo targets are complementary, while the final sample weight provides a smaller additional gain.

\noindent\textbf{Pseudo-target reliability.}
\cref{tab:pseudo_reliability_summary} evaluates low-clean+noisy samples, where refurbishment depends most on the pseudo branch. TRACE improves every reported setting, averaging +13.32\% across the three CIFAR-100 symmetric-noise rates. CIFAR-10 and CIFAR-10N show the same direction under a smaller label space and human annotation noise, whereas the smaller +0.70-point gain on CIFAR-100N marks the harder real-noise boundary. \cref{tab:appendix_cross_dataset_reliability} reports the full matrix.

\begin{figure}[t]
  \centering
  \includegraphics[width=\columnwidth]{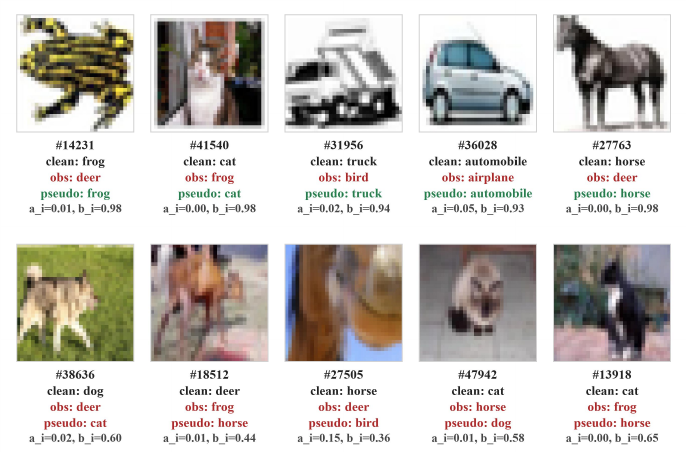}
  \caption{CIFAR-10 examples under 50\% symmetric noise. \(a_i\) and \(b_i\) are the observed-label and pseudo-target branch weights in \cref{eq:weights}. Top: confident pseudo targets are used for correction. Bottom: unreliable pseudo targets are down-weighted. Zoom in to see details.}
  \label{fig:case_study_cifar10}
  \vspace{-1em}
\end{figure}

\begin{figure}[t]
  \centering
  \begin{subfigure}[t]{0.49\columnwidth}
    \centering
    \includegraphics[width=\linewidth]{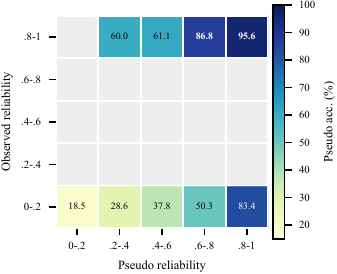}
    \caption{RoLR}
    \label{fig:reliability_landscape_a}
  \end{subfigure}
  \hfill
  \begin{subfigure}[t]{0.49\columnwidth}
    \centering
    \includegraphics[width=\linewidth]{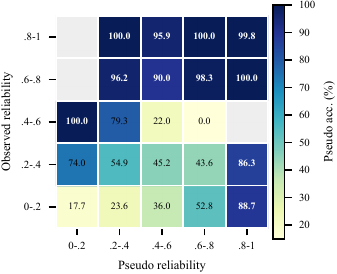}
    \caption{TRACE}
    \label{fig:reliability_landscape_b}
  \end{subfigure}
  \caption{Reliability landscapes on CIFAR-100 with 50\% sym. noise. Cells show pseudo-target accuracy (\%); cells containing less than 0.01\% of samples are left blank. The observed axis uses loss-based cleanliness for RoLR and refined reliability for TRACE.}
  \label{fig:reliability_landscape}
  \vspace{-1em}
\end{figure}

\noindent\textbf{Error quality on CIFAR-100.}
\cref{fig:exp_pseudo_reliability_a,fig:exp_pseudo_reliability_b} show how reliability varies across cleanliness bins on CIFAR-100 with 50\% symmetric noise. TRACE raises pseudo-target accuracy from 70.75\% to 74.15\% on low-clean samples and from 72.92\% to 84.70\% on low-clean+noisy samples, where the pseudo branch matters most.
\cref{fig:exp_pseudo_reliability_c,fig:exp_pseudo_reliability_d} further show fewer high-confidence errors (8.93\% to 6.16\%) and lower ECE (13.03\% to 8.52\%). These results indicate better pseudo-target quality rather than merely more pseudo-target selection.
These two diagnostics matter because a confidently wrong pseudo target can receive a large training weight and repeatedly reinforce an noisy class; accuracy alone does not expose that risk.

\noindent\textbf{Case study.}
\cref{fig:case_study_cifar10} shows the same behavior at sample level. In the top row, a noisy label gives small \(a_i\), while a correct, confident pseudo target gives large \(b_i\). In the bottom row, TRACE keeps \(b_i\) small because the replacement is wrong or uncertain. Thus, the need for correction doesn't override pseudo-target reliability.
The cases operationalize the two decisions: \(a_i\) asks whether to retain the observed label, while \(b_i\) asks whether the proposed replacement is usable.

\noindent\textbf{Reliability landscape.}
\cref{fig:reliability_landscape} jointly stratifies samples by observed-label and pseudo-target reliability, rather than examining either score marginally. Low observed-label reliability alone does not justify the pseudo branch: accuracy remains poor in low-pseudo bins and rises in the high-pseudo bin. In the key low-observed/high-pseudo region, TRACE raises pseudo-target accuracy from 83.4\% to 88.7\% and reduces high-confidence errors from 12.3\% to 8.3\%, supporting separate reliability estimates for the two sources.

\noindent\textbf{Sensitivity to hyperparameters.}
\cref{tab:sensitivity_hyperparameters} varies the loss--structure balance \(\alpha\) and pseudo-confidence power \(\rho\) one at a time on CIFAR-100 with 50\% symmetric noise. The default \(\alpha=0.7\) gives the best accuracy, pseudo-target accuracy, and high-confidence error rate; \(\rho=1.0\) gives the best accuracy and pseudo-target accuracy, while \(\rho=1.5\) further reduces high-confidence errors with small losses elsewhere. Additional factors are reported in \cref{tab:appendix_sensitivity_hyperparameters}.
We therefore use \(\rho=1.0\) as the balanced default rather than optimizing only the high-confidence-error metric.

\section{Conclusion}


This work identifies the risk of replacing noise with noise in noisy-label refurbishment: low trust in an observed label does not establish that its pseudo target is reliable. TRACE addresses this mismatch through separate reliability assessment for the observed label and pseudo target. It combines loss, shallow-anchored relation stability, and prediction agreement for the former, while an independent confidence signal assesses the latter; the resulting source-specific weights control label correction and sample reweighting. Experiments across synthetic, human-annotated, and real-world noise show that TRACE improves refurbishment learners and pseudo-target reliability. Findings support source-specific reliability assessment as a safer interface for noisy-label learning.


\bibliography{aaai2027}

@String(CVPR= {IEEE Conf. Comput. Vis. Pattern Recog.})

@String(ICCV= {Int. Conf. Comput. Vis.})

@String(ICLR = {Int. Conf. Learn. Represent.})

@String(IJCAI = {IJCAI})

@String(AAAI = {AAAI})

@String(CVPR  = {CVPR})

@String(ICCV  = {ICCV})

@String(ICLR  = {ICLR})

@String{Computing = "Computing" }

@String{Computer = "{IEEE} Computer" }

@String{Springer = "Springer-Verlag" }

@BOOK{test,
   author = "Donald E. Knuth",
   title = "Seminumerical Algorithms",
   volume = 2,
   series = "The Art of Computer Programming",
   publisher = "Addison-Wesley",
   address = "Reading, MA",
   edition = "2nd",
   month = "10~" # jan,
   year = "1981",
}

@ArtifactSoftware{R,
    title = {R: A Language and Environment for Statistical Computing},
    author = {{R Core Team}},
    organization = {R Foundation for Statistical Computing},
    address = {Vienna, Austria},
    year = {2019},
    url = {https://www.R-project.org/},
}

@inproceedings{DBLP:conf/mm/TanXWL21,
  author    = {Cheng Tan and
               Jun Xia and
               Lirong Wu and
               Stan Z. Li},
  editor    = {Heng Tao Shen and
               Yueting Zhuang and
               John R. Smith and
               Yang Yang and
               Pablo Cesar and
               Florian Metze and
               Balakrishnan Prabhakaran},
  title     = {Co-learning: Learning from Noisy Labels with Self-supervision},
  booktitle = {{MM} '21: {ACM} Multimedia Conference, Virtual Event, China, October
               20 - 24, 2021},
  pages     = {1405--1413},
  publisher = {{ACM}},
  year      = {2021},
  url       = {https://doi.org/10.1145/3474085.3475622},
  doi       = {10.1145/3474085.3475622},
  bibsource = {dblp computer science bibliography, https://dblp.org}
}

@inproceedings{DBLP:conf/nips/NatarajanDRT13,
  author    = {Nagarajan Natarajan and
               Inderjit S. Dhillon and
               Pradeep Ravikumar and
               Ambuj Tewari},
  editor    = {Christopher J. C. Burges and
               L{\'{e}}on Bottou and
               Zoubin Ghahramani and
               Kilian Q. Weinberger},
  title     = {Learning with Noisy Labels},
  booktitle = {Advances in Neural Information Processing Systems 26: 27th Annual
               Conference on Neural Information Processing Systems 2013. Proceedings
               of a meeting held December 5-8, 2013, Lake Tahoe, Nevada, United States},
  pages     = {1196--1204},
  year      = {2013},
  url       = {https://proceedings.neurips.cc/paper/2013/hash/3871bd64012152bfb53fdf04b401193f-Abstract.html},
  bibsource = {dblp computer science bibliography, https://dblp.org}
}

@article{krizhevsky2009learning,
  title={Learning multiple layers of features from tiny images},
  author={Krizhevsky, Alex and Hinton, Geoffrey and others},
  year={2009},
  publisher={Citeseer}
}

@article{Wang2018d,
archivePrefix = {arXiv},
arxivId = {1804.00092},
author = {Wang, Yisen and Liu, Weiyang and Ma, Xingjun and Bailey, James and Zha, Hongyuan and Song, Le and Xia, Shu Tao},
doi = {10.1109/CVPR.2018.00906},
eprint = {1804.00092},
isbn = {9781538664209},
issn = {10636919},
journal = {Proceedings of the IEEE Computer Society Conference on Computer Vision and Pattern Recognition},
pages = {8688--8696},
title = {{Iterative Learning with Open-set Noisy Labels}},
year = {2018}
}

@inproceedings{DBLP:conf/cvpr/PatriniRMNQ17,
  author    = {Giorgio Patrini and
               Alessandro Rozza and
               Aditya Krishna Menon and
               Richard Nock and
               Lizhen Qu},
  title     = {Making Deep Neural Networks Robust to Label Noise: {A} Loss Correction
               Approach},
  booktitle = {2017 {IEEE} Conference on Computer Vision and Pattern Recognition,
               {CVPR} 2017, Honolulu, HI, USA, July 21-26, 2017},
  pages     = {2233--2241},
  publisher = {{IEEE} Computer Society},
  year      = {2017},
  url       = {https://doi.org/10.1109/CVPR.2017.240},
  doi       = {10.1109/CVPR.2017.240},
  bibsource = {dblp computer science bibliography, https://dblp.org}
}

@inproceedings{Song2019,
  author    = {Hwanjun Song and
               Minseok Kim and
               Jae{-}Gil Lee},
  editor    = {Kamalika Chaudhuri and
               Ruslan Salakhutdinov},
  title     = {{SELFIE:} Refurbishing Unclean Samples for Robust Deep Learning},
  series    = {Proceedings of Machine Learning Research},
  volume    = {97},
  pages     = {5907--5915},
  publisher = {{PMLR}},
  year      = {2019},
  url       = {http://proceedings.mlr.press/v97/song19b.html},
  bibsource = {dblp computer science bibliography, https://dblp.org}
}

@inproceedings{DBLP:conf/cvpr/WeiFC020,
  author    = {Hongxin Wei and
               Lei Feng and
               Xiangyu Chen and
               Bo An},
  title     = {Combating Noisy Labels by Agreement: {A} Joint Training Method with
               Co-Regularization},
  booktitle = {{CVPR} 2020, Seattle, WA, USA, June 13-19, 2020},
  pages     = {13723--13732},
  publisher = {Computer Vision Foundation / {IEEE}},
  year      = {2020},
  url       = {https://openaccess.thecvf.com/content\_CVPR\_2020/html/Wei\_Combating\_Noisy\_Labels\_by\_Agreement\_A\_Joint\_Training\_Method\_with\_CVPR\_2020\_paper.html},
  doi       = {10.1109/CVPR42600.2020.01374},
  bibsource = {dblp computer science bibliography, https://dblp.org}
}

@article{Jiang2017,
archivePrefix = {arXiv},
arxivId = {1712.05055},
author = {Jiang, Lu and Zhou, Zhengyuan and Leung, Thomas and Li, Li Jia and Fei-Fei, Li},
eprint = {1712.05055},
isbn = {9781510867963},
journal = {35th International Conference on Machine Learning, ICML 2018},
number = {6},
pages = {3601--3620},
title = {{Mentornet: Learning data-driven curriculum for very deep neural networks on corrupted labels}},
volume = {5},
year = {2018}
}

@article{Malach2017,
archivePrefix = {arXiv},
arxivId = {1706.02613},
author = {Malach, Eran and Shalev-Shwartz, Shai},
eprint = {1706.02613},
issn = {10495258},
journal = {Advances in Neural Information Processing Systems},
month = {jun},
pages = {961--971},
title = {{Decoupling "when to update" from "how to update"}},
url = {http://arxiv.org/abs/1706.02613},
volume = {2017-Decem},
year = {2017}
}

@article{Han2018c,
archivePrefix = {arXiv},
arxivId = {1804.06872},
author = {Han, Bo and Yao, Quanming and Yu, Xingrui and Niu, Gang and Xu, Miao and Hu, Weihua and Tsang, Ivor W. and Sugiyama, Masashi},
eprint = {1804.06872},
issn = {10495258},
journal = {Advances in Neural Information Processing Systems},
number = {NeurIPS},
pages = {8527--8537},
title = {{Co-teaching: Robust training of deep neural networks with extremely noisy labels}},
volume = {2018-Decem},
year = {2018}
}

@inproceedings{DBLP:conf/aaai/GhoshKS17,
  author    = {Aritra Ghosh and
               Himanshu Kumar and
               P. S. Sastry},
  title     = {Robust Loss Functions under Label Noise for Deep Neural Networks},
  booktitle = {Proceedings of the Thirty-First {AAAI} Conference on Artificial Intelligence},
  pages     = {1919--1925},
  publisher = {{AAAI} Press},
  year      = {2017},
  url       = {http://aaai.org/ocs/index.php/AAAI/AAAI17/paper/view/14759},
  bibsource = {dblp computer science bibliography, https://dblp.org}
}

@inproceedings{DBLP:conf/icml/ZhengWG0MC20,
  author       = {Songzhu Zheng and
                  Pengxiang Wu and
                  Aman Goswami and
                  Mayank Goswami and
                  Dimitris N. Metaxas and
                  Chao Chen},
  title        = {Error-Bounded Correction of Noisy Labels},
  booktitle    = {Proceedings of the 37th International Conference on Machine Learning,
                  {ICML} 2020, 13-18 July 2020, Virtual Event},
  series       = {Proceedings of Machine Learning Research},
  volume       = {119},
  pages        = {11447--11457},
  publisher    = {{PMLR}},
  year         = {2020},
  url          = {http://proceedings.mlr.press/v119/zheng20c.html},
  bibsource    = {dblp computer science bibliography, https://dblp.org}
}

@inproceedings{DBLP:conf/iclr/LiSH20,
  author       = {Junnan Li and
                  Richard Socher and
                  Steven C. H. Hoi},
  title        = {DivideMix: Learning with Noisy Labels as Semi-supervised Learning},
  booktitle    = {8th International Conference on Learning Representations, {ICLR} 2020,
                  Addis Ababa, Ethiopia, April 26-30, 2020},
  publisher    = {OpenReview.net},
  year         = {2020},
  url          = {https://openreview.net/forum?id=HJgExaVtwr},
  bibsource    = {dblp computer science bibliography, https://dblp.org}
}

@inproceedings{DBLP:conf/aaai/WuSX0M21,
  author       = {Yichen Wu and
                  Jun Shu and
                  Qi Xie and
                  Qian Zhao and
                  Deyu Meng},
  title        = {Learning to Purify Noisy Labels via Meta Soft Label Corrector},
  booktitle    = {Thirty-Fifth {AAAI} Conference on Artificial Intelligence, {AAAI}
                  2021},
  pages        = {10388--10396},
  publisher    = {{AAAI} Press},
  year         = {2021},
  url          = {https://ojs.aaai.org/index.php/AAAI/article/view/17244},
  bibsource    = {dblp computer science bibliography, https://dblp.org}
}

@inproceedings{DBLP:conf/cvpr/LiWZK19,
  author       = {Junnan Li and
                  Yongkang Wong and
                  Qi Zhao and
                  Mohan S. Kankanhalli},
  title        = {Learning to Learn From Noisy Labeled Data},
  booktitle    = {{IEEE} Conference on Computer Vision and Pattern Recognition, {CVPR}
                  2019, Long Beach, CA, USA, June 16-20, 2019},
  pages        = {5051--5059},
  publisher    = {Computer Vision Foundation / {IEEE}},
  year         = {2019},
  url          = {http://openaccess.thecvf.com/content\_CVPR\_2019/html/Li\_Learning\_to\_Learn\_From\_Noisy\_Labeled\_Data\_CVPR\_2019\_paper.html},
  doi          = {10.1109/CVPR.2019.00519},
  bibsource    = {dblp computer science bibliography, https://dblp.org}
}

@inproceedings{DBLP:conf/iclr/WeiZ0L0022,
  author       = {Jiaheng Wei and
                  Zhaowei Zhu and
                  Hao Cheng and
                  Tongliang Liu and
                  Gang Niu and
                  Yang Liu},
  title        = {Learning with Noisy Labels Revisited: {A} Study Using Real-World Human
                  Annotations},
  booktitle    = {{ICLR}
                  2022},
  publisher    = {OpenReview.net},
  year         = {2022},
  url          = {https://openreview.net/forum?id=TBWA6PLJZQm},
  bibsource    = {dblp computer science bibliography, https://dblp.org}
}

@article{Iscen2022LearningWN,
  title={Learning with Neighbor Consistency for Noisy Labels},
  author={Ahmet Iscen and Jack Valmadre and Anurag Arnab and Cordelia Schmid},
  journal={2022 IEEE/CVF Conference on Computer Vision and Pattern Recognition (CVPR)},
  year={2022},
  pages={4662-4671},
  url={https://api.semanticscholar.org/CorpusID:246607828}
}

@article{Yi2022OnLC,
  title={On Learning Contrastive Representations for Learning with Noisy Labels},
  author={Linya Yi and Sheng Liu and Qi She and Alex McLeod and Boyu Wang},
  journal={2022 IEEE/CVF Conference on Computer Vision and Pattern Recognition (CVPR)},
  year={2022},
  pages={16661-16670},
  url={https://api.semanticscholar.org/CorpusID:247223119}
}

@inproceedings{Chen2020BeyondCA,
  title={Beyond Class-Conditional Assumption: A Primary Attempt to Combat Instance-Dependent Label Noise},
  author={Pengfei Chen and Junjie Ye and Guangyong Chen and Jingwei Zhao and Pheng-Ann Heng},
  booktitle={AAAI Conference on Artificial Intelligence},
  year={2020},
  url={https://api.semanticscholar.org/CorpusID:228083495}
}

@inproceedings{Wei2022MitigatingMO,
  title={Mitigating Memorization of Noisy Labels by Clipping the Model Prediction},
  author={Hongxin Wei and Huiping Zhuang and Renchunzi Xie and Lei Feng and Gang Niu and Bo An and Yixuan Li},
booktitle={{ICML} 2023},
  year={2023},
  url={https://api.semanticscholar.org/CorpusID:259138834}
}

@inproceedings{Chen2021TwoWD,
  author       = {Mingcai Chen and
                  Hao Cheng and
                  Yuntao Du and
                  Ming Xu and
                  Wenyu Jiang and
                  Chongjun Wang},
  editor       = {Brian Williams and
                  Yiling Chen and
                  Jennifer Neville},
  title        = {Two Wrongs Don't Make a Right: Combating Confirmation Bias in Learning
                  with Label Noise},
  booktitle    = {Thirty-Seventh {AAAI} Conference on Artificial Intelligence, {AAAI}
                  2023, Thirty-Fifth Conference on Innovative Applications of Artificial
                  Intelligence, {IAAI} 2023, Thirteenth Symposium on Educational Advances
                  in Artificial Intelligence, {EAAI} 2023, Washington, DC, USA, February
                  7-14, 2023},
  pages        = {14765--14773},
  publisher    = {{AAAI} Press},
  year         = {2023},
  url          = {https://doi.org/10.1609/aaai.v37i12.26725},
  doi          = {10.1609/AAAI.V37I12.26725},
  bibsource    = {dblp computer science bibliography, https://dblp.org}
}

@inproceedings{DBLP:conf/iccv/Zhang0FLCLL23,
  author       = {Ziyi Zhang and
                  Weikai Chen and
                  Chaowei Fang and
                  Zhen Li and
                  Lechao Chen and
                  Liang Lin and
                  Guanbin Li},
  title        = {RankMatch: Fostering Confidence and Consistency in Learning with Noisy
                  Labels},
  booktitle    = {{IEEE/CVF} International Conference on Computer Vision, {ICCV} 2023,
                  Paris, France, October 1-6, 2023},
  pages        = {1644--1654},
  publisher    = {{IEEE}},
  year         = {2023},
  url          = {https://doi.org/10.1109/ICCV51070.2023.00158},
  doi          = {10.1109/ICCV51070.2023.00158},
  bibsource    = {dblp computer science bibliography, https://dblp.org}
}

@article{DBLP:journals/corr/abs-1708-02862,
  author       = {Wen Li and
                  Limin Wang and
                  Wei Li and
                  Eirikur Agustsson and
                  Luc Van Gool},
  title        = {WebVision Database: Visual Learning and Understanding from Web Data},
  journal      = {CoRR},
  volume       = {abs/1708.02862},
  year         = {2017},
  url          = {http://arxiv.org/abs/1708.02862},
  eprinttype    = {arXiv},
  eprint       = {1708.02862},
  bibsource    = {dblp computer science bibliography, https://dblp.org}
}

@inproceedings{DBLP:conf/icml/KimBZL23,
  author       = {Jihye Kim and
                  Aristide Baratin and
                  Yan Zhang and
                  Simon Lacoste{-}Julien},
  editor       = {Andreas Krause and
                  Emma Brunskill and
                  Kyunghyun Cho and
                  Barbara Engelhardt and
                  Sivan Sabato and
                  Jonathan Scarlett},
  title        = {CrossSplit: Mitigating Label Noise Memorization through Data Splitting},
  booktitle    = {International Conference on Machine Learning, {ICML} 2023, 23-29 July
                  2023, Honolulu, Hawaii, {USA}},
  series       = {Proceedings of Machine Learning Research},
  volume       = {202},
  pages        = {16377--16392},
  publisher    = {{PMLR}},
  year         = {2023},
  url          = {https://proceedings.mlr.press/v202/kim23a.html},
  bibsource    = {dblp computer science bibliography, https://dblp.org}
}

@inproceedings{DBLP:conf/cvpr/Li0S023,
  author       = {Yifan Li and
                  Hu Han and
                  Shiguang Shan and
                  Xilin Chen},
  title        = {{DISC:} Learning from Noisy Labels via Dynamic Instance-Specific Selection
                  and Correction},
  booktitle    = {{IEEE/CVF} Conference on Computer Vision and Pattern Recognition,
                  {CVPR} 2023, Vancouver, BC, Canada, June 17-24, 2023},
  pages        = {24070--24079},
  publisher    = {{IEEE}},
  year         = {2023},
  url          = {https://doi.org/10.1109/CVPR52729.2023.02305},
  doi          = {10.1109/CVPR52729.2023.02305},
  bibsource    = {dblp computer science bibliography, https://dblp.org}
}

@inproceedings{DBLP:conf/cvpr/TuZLLLWWZ23,
  author       = {Yuanpeng Tu and
                  Boshen Zhang and
                  Yuxi Li and
                  Liang Liu and
                  Jian Li and
                  Yabiao Wang and
                  Chengjie Wang and
                  Cairong Zhao},
  title        = {Learning from Noisy Labels with Decoupled Meta Label Purifier},
  booktitle    = {{IEEE/CVF} Conference on Computer Vision and Pattern Recognition,
                  {CVPR} 2023, Vancouver, BC, Canada, June 17-24, 2023},
  pages        = {19934--19943},
  publisher    = {{IEEE}},
  year         = {2023},
  url          = {https://doi.org/10.1109/CVPR52729.2023.01909},
  doi          = {10.1109/CVPR52729.2023.01909},
  bibsource    = {dblp computer science bibliography, https://dblp.org}
}

@inproceedings{DBLP:conf/cvpr/KarimRRMS22,
  author       = {Nazmul Karim and
                  Mamshad Nayeem Rizve and
                  Nazanin Rahnavard and
                  Ajmal Mian and
                  Mubarak Shah},
  title        = {{UNICON:} Combating Label Noise Through Uniform Selection and Contrastive
                  Learning},
  booktitle    = {{IEEE/CVF} Conference on Computer Vision and Pattern Recognition,
                  {CVPR} 2022, New Orleans, LA, USA, June 18-24, 2022},
  pages        = {9666--9676},
  publisher    = {{IEEE}},
  year         = {2022},
  url          = {https://doi.org/10.1109/CVPR52688.2022.00945},
  doi          = {10.1109/CVPR52688.2022.00945},
  bibsource    = {dblp computer science bibliography, https://dblp.org}
}

@inproceedings{DBLP:conf/cvpr/LiXGL22,
  author       = {Shikun Li and
                  Xiaobo Xia and
                  Shiming Ge and
                  Tongliang Liu},
  title        = {Selective-Supervised Contrastive Learning with Noisy Labels},
  booktitle    = {{IEEE/CVF} Conference on Computer Vision and Pattern Recognition,
                  {CVPR} 2022, New Orleans, LA, USA, June 18-24, 2022},
  pages        = {316--325},
  publisher    = {{IEEE}},
  year         = {2022},
  url          = {https://doi.org/10.1109/CVPR52688.2022.00041},
  doi          = {10.1109/CVPR52688.2022.00041},
  bibsource    = {dblp computer science bibliography, https://dblp.org}
}

@inproceedings{DBLP:conf/cvpr/XiaoXYHW15,
  author       = {Tong Xiao and
                  Tian Xia and
                  Yi Yang and
                  Chang Huang and
                  Xiaogang Wang},
  title        = {Learning from massive noisy labeled data for image classification},
  booktitle    = {{IEEE} Conference on Computer Vision and Pattern Recognition, {CVPR}
                  2015, Boston, MA, USA, June 7-12, 2015},
  pages        = {2691--2699},
  publisher    = {{IEEE} Computer Society},
  year         = {2015},
  url          = {https://doi.org/10.1109/CVPR.2015.7298885},
  doi          = {10.1109/CVPR.2015.7298885},
  bibsource    = {dblp computer science bibliography, https://dblp.org}
}

@article{DBLP:journals/ijcv/KimRCK25,
  author       = {Daehwan Kim and
                  Kwangrok Ryoo and
                  Hansang Cho and
                  Seungryong Kim},
  title        = {SplitNet: Learnable Clean-Noisy Label Splitting for Learning with
                  Noisy Labels},
  journal      = {Int. J. Comput. Vis.},
  volume       = {133},
  number       = {2},
  pages        = {549--566},
  year         = {2025},
  url          = {https://doi.org/10.1007/s11263-024-02187-4},
  doi          = {10.1007/S11263-024-02187-4},
  bibsource    = {dblp computer science bibliography, https://dblp.org}
}

@ARTICLE{Wang2024-nj,
  title     = "Using unreliable pseudo-labels for label-efficient semantic segmentation",
  author    = "Wang, Haochen and Wang, Yuchao and Shen, Yujun and Fan, Junsong
               and Wang, Yuxi and Zhang, Zhaoxiang",
  journal   = "Int. J. Comput. Vis.",
  publisher = "Springer Science and Business Media LLC",
  month     =  oct,
  year      =  2024,
  copyright = "https://www.springernature.com/gp/researchers/text-and-data-mining",
  language  = "en"
}

@article{DBLP:journals/corr/abs-2410-07689,
  author       = {Keryan Chelouche and
                  Marie Lachaize and
                  Marine Bernard and
                  Louise Olgiati and
                  R{\'{e}}mi Cuingnet},
  title        = {When the Small-Loss Trick is Not Enough: Multi-Label Image Classification
                  with Noisy Labels Applied to {CCTV} Sewer Inspections},
  journal      = {CoRR},
  volume       = {abs/2410.07689},
  year         = {2024},
  url          = {https://doi.org/10.48550/arXiv.2410.07689},
  doi          = {10.48550/ARXIV.2410.07689},
  eprinttype    = {arXiv},
  eprint       = {2410.07689},
  bibsource    = {dblp computer science bibliography, https://dblp.org}
}

@article{DBLP:journals/corr/abs-2403-06869,
  author       = {Hao Chen and
                  Jindong Wang and
                  Zihan Wang and
                  Ran Tao and
                  Hongxin Wei and
                  Xing Xie and
                  Masashi Sugiyama and
                  Bhiksha Raj},
  title        = {Learning with Noisy Foundation Models},
  journal      = {CoRR},
  volume       = {abs/2403.06869},
  year         = {2024},
  url          = {https://doi.org/10.48550/arXiv.2403.06869},
  doi          = {10.48550/ARXIV.2403.06869},
  eprinttype    = {arXiv},
  eprint       = {2403.06869},
  bibsource    = {dblp computer science bibliography, https://dblp.org}
}

@inproceedings{DBLP:conf/aaai/FanL25,
  author       = {Wenxiao Fan and
                  Kan Li},
  editor       = {Toby Walsh and
                  Julie Shah and
                  Zico Kolter},
  title        = {Combating Semantic Contamination in Learning with Label Noise},
  booktitle    = {AAAI-25, Sponsored by the Association for the Advancement of Artificial
                  Intelligence, February 25 - March 4, 2025, Philadelphia, PA, {USA}},
  pages        = {2870--2878},
  publisher    = {{AAAI} Press},
  year         = {2025},
  url          = {https://doi.org/10.1609/aaai.v39i3.32293},
  doi          = {10.1609/AAAI.V39I3.32293},
  bibsource    = {dblp computer science bibliography, https://dblp.org}
}

@article{DBLP:journals/pr/CordeiroC25,
  author       = {Filipe R. Cordeiro and
                  Gustavo Carneiro},
  title        = {{ANNE:} Adaptive Nearest Neighbours and Eigenvector-based sample selection for robust learning with noisy labels},
  journal      = {Pattern Recognit.},
  volume       = {159},
  pages        = {111132},
  year         = {2025},
  url          = {https://doi.org/10.1016/j.patcog.2024.111132},
  doi          = {10.1016/J.PATCOG.2024.111132},
  bibsource    = {dblp computer science bibliography, https://dblp.org}
}

@inproceedings{DBLP:conf/ijcai/LuH22,
  author       = {Yangdi Lu and
                  Wenbo He},
  editor       = {Luc De Raedt},
  title        = {{SELC:} Self-Ensemble Label Correction Improves Learning with Noisy
                  Labels},
  booktitle    = {Proceedings of the Thirty-First International Joint Conference on
                  Artificial Intelligence, {IJCAI} 2022, Vienna, Austria, 23-29 July
                  2022},
  pages        = {3278--3284},
  publisher    = {ijcai.org},
  year         = {2022},
  url          = {https://doi.org/10.24963/ijcai.2022/455},
  doi          = {10.24963/IJCAI.2022/455},
  bibsource    = {dblp computer science bibliography, https://dblp.org}
}

@inproceedings{DBLP:conf/aistats/WongsoGM23,
  author       = {Shelvia Wongso and
                  Rohan Ghosh and
                  Mehul Motani},
  editor       = {Francisco J. R. Ruiz and
                  Jennifer G. Dy and
                  Jan{-}Willem van de Meent},
  title        = {Using Sliced Mutual Information to Study Memorization and Generalization
                  in Deep Neural Networks},
  booktitle    = {International Conference on Artificial Intelligence and Statistics,
                  25-27 April 2023, Palau de Congressos, Valencia, Spain},
  series       = {Proceedings of Machine Learning Research},
  pages        = {11608--11629},
  publisher    = {{PMLR}},
  year         = {2023},
  url          = {https://proceedings.mlr.press/v206/wongso23a.html},
  bibsource    = {dblp computer science bibliography, https://dblp.org}
}

@inproceedings{DBLP:conf/icml/MainiMSLKZ23,
  author       = {Pratyush Maini and
                  Michael Curtis Mozer and
                  Hanie Sedghi and
                  Zachary Chase Lipton and
                  J. Zico Kolter and
                  Chiyuan Zhang},
  editor       = {Andreas Krause and
                  Emma Brunskill and
                  Kyunghyun Cho and
                  Barbara Engelhardt and
                  Sivan Sabato and
                  Jonathan Scarlett},
  title        = {Can Neural Network Memorization Be Localized?},
  booktitle    = {International Conference on Machine Learning, {ICML} 2023, 23-29 July
                  2023, Honolulu, Hawaii, {USA}},
  series       = {Proceedings of Machine Learning Research},
  pages        = {23536--23557},
  publisher    = {{PMLR}},
  year         = {2023},
  url          = {https://proceedings.mlr.press/v202/maini23a.html},
  bibsource    = {dblp computer science bibliography, https://dblp.org}
}

@article{DBLP:journals/corr/abs-2008-08186,
  author       = {Vardan Papyan and
                  X. Y. Han and
                  David L. Donoho},
  title        = {Prevalence of Neural Collapse during the terminal phase of deep learning
                  training},
  journal      = {CoRR},
  volume       = {abs/2008.08186},
  year         = {2020},
  url          = {https://arxiv.org/abs/2008.08186},
  eprinttype   = {arXiv},
  eprint       = {2008.08186},
  bibsource    = {dblp computer science bibliography, https://dblp.org}
}

@inproceedings{DBLP:conf/aaai/FanL26,
  author       = {Wenxiao Fan and
                  Kan Li},
  editor       = {Sven Koenig and
                  Chad Jenkins and
                  Matthew E. Taylor},
  title        = {Leveraging Dissimilarity Invariance as a Robust Anchor for Learning
                  with Noisy Labels},
  booktitle    = {Fortieth {AAAI} Conference on Artificial Intelligence, Thirty-Eighth
                  Conference on Innovative Applications of Artificial Intelligence,
                  Sixteenth Symposium on Educational Advances in Artificial Intelligence,
                  {AAAI} 2026, Singapore, January 20-27, 2026},
  pages        = {3804--3812},
  publisher    = {{AAAI} Press},
  year         = {2026},
  url          = {https://doi.org/10.1609/aaai.v40i5.37381},
  doi          = {10.1609/AAAI.V40I5.37381},
  bibsource    = {dblp computer science bibliography, https://dblp.org}
}

@inproceedings{DBLP:conf/aaai/Pan00D25,
  author       = {Weiran Pan and
                  Wei Wei and
                  Feida Zhu and
                  Yong Deng},
  editor       = {Toby Walsh and
                  Julie Shah and
                  Zico Kolter},
  title        = {Enhanced Sample Selection with Confidence Tracking: Identifying Correctly
                  Labeled Yet Hard-to-Learn Samples in Noisy Data},
  booktitle    = {Thirty-Ninth {AAAI} Conference on Artificial Intelligence, Thirty-Seventh
                  Conference on Innovative Applications of Artificial Intelligence,
                  Fifteenth Symposium on Educational Advances in Artificial Intelligence,
                  {AAAI} 2025, Philadelphia, PA, USA, February 25 - March 4, 2025},
  pages        = {19795--19803},
  publisher    = {{AAAI} Press},
  year         = {2025},
  url          = {https://doi.org/10.1609/aaai.v39i19.34180},
  doi          = {10.1609/AAAI.V39I19.34180},
  bibsource    = {dblp computer science bibliography, https://dblp.org}
}

@inproceedings{DBLP:conf/nips/KimKCCY21,
  author       = {Taehyeon Kim and
                  Jongwoo Ko and
                  Sangwook Cho and
                  Jinhwan Choi and
                  Se{-}Young Yun},
  editor       = {Marc'Aurelio Ranzato and
                  Alina Beygelzimer and
                  Yann N. Dauphin and
                  Percy Liang and
                  Jennifer Wortman Vaughan},
  title        = {{FINE} Samples for Learning with Noisy Labels},
  booktitle    = {Advances in Neural Information Processing Systems 34: Annual Conference
                  on Neural Information Processing Systems 2021, NeurIPS 2021, December
                  6-14, 2021, virtual},
  pages        = {24137--24149},
  year         = {2021},
  url          = {https://proceedings.neurips.cc/paper/2021/hash/ca91c5464e73d3066825362c3093a45f-Abstract.html},
  bibsource    = {dblp computer science bibliography, https://dblp.org}
}

@inproceedings{lee2018cleannet,
  title        = {CleanNet: Transfer Learning for Scalable Image Classifier Training with Label Noise},
  author       = {Lee, Kuang-Huei and He, Xiaodong and Zhang, Lei and Yang, Linjun},
  booktitle    = {Proceedings of the IEEE Conference on Computer Vision and Pattern Recognition},
  pages        = {5447--5456},
  year         = {2018}
}

\clearpage
\appendix

\noindent\textbf{Appendix Overview.}
\begin{itemize}
  \item Appendix~A reports default hyperparameters, compute resources, and asset information.
  \item Appendix~B develops the two-source reliability and shallow-anchor analyses, clarifies the treatment of hard-but-clean samples, and provides extended representation diagnostics.
  \item Appendix~C specifies the experimental protocols, ablation variants, and synthetic-noise construction.
  \item Appendix~D reports additional noise settings, sensitivity analyses, reliability statistics, and real-noise diagnostics.
  \item Appendix~E explains the construction and interpretation of the diagnostic figures.
  \item Appendix~F presents TRACE pseudocode, plug-in implementation details, and computational overhead.
  \item Appendix~G discusses scope limitations and the extension to open-set label noise.
\end{itemize}

\paragraph{AI Use Disclosure.}
Generative AI tools were used to assist with language editing, code
debugging, and figure drafting. The authors reviewed and verified all
AI-assisted outputs and take full responsibility for the content of
this manuscript.

\section{Default Hyperparameters and Resources}
\label{app:default_hyperparameters}

\cref{tab:default_hyperparameters} reports the default hyperparameters used for CIFAR-style TRACE experiments with the RoLR instantiation. The dataset, noise type, noise rate, and base learner are changed according to the experimental setting; all other values are kept fixed unless an ablation explicitly states otherwise. For large-scale real-noise experiments, we follow the training protocol of the corresponding base learner and use the same reliability defaults when TRACE is attached.

\begin{table*}[htbp]
  \centering
  \footnotesize
  \setlength{\tabcolsep}{4pt}
  \begin{tabular}{p{0.20\textwidth}p{0.28\textwidth}p{0.38\textwidth}}
    \toprule
    \textbf{Category} & \textbf{Hyperparameter} & \textbf{Default value} \\
    \midrule
    Architecture & Backbone & ResNet-18 \\
    Training & Batch size & 64 \\
    Training & Epochs / warm-up & 500 / 15 \\
    Training & Optimizer & SGD, momentum \(0.9\), weight decay \(5\times10^{-4}\) \\
    Training & Initial learning rate & \(0.02\), divided by 10 in the last 100 epochs \\
    Training & Supervised / prior penalty weights & \(\lambda_l=1,\ \lambda_p=1\) \\
    Pseudo target & Sharpening temperature & \(T=1\) \\
    Loss posterior & GMM components & 2 \\
    Loss posterior & GMM max iterations / tolerance / regularizer & \(10 / 10^{-2} / 5\times10^{-4}\) \\
    Observed label & Confidence mode & loss+structure \\
    Observed label & Loss-structure fusion weight & \(\alpha=0.7\) \\
    Structure confidence & Transport drift temperature & \(\gamma=5.0\) \\
    Structure confidence & Sparse relation neighbors & \(k=50\) \\
    Structure confidence & Start epoch / ramp-up length & \(30 / 20\) \\
    Agreement gate & Disagreement multiplier & \(\lambda_{\mathrm{dis}}=0.5\) \\
    Pseudo-target reliability & Start epoch / confidence power & \(30 / 1.0\) \\
    Supervision strength & Minimum sample weight & \(w_{\min}=0.2\) \\
    \bottomrule
  \end{tabular}
   \caption{Default hyperparameters for the CIFAR-style RoLR+TRACE instantiation.}
  \label{tab:default_hyperparameters}
\end{table*}

\begin{table}[htbp]
  \centering
  \small
  \setlength{\tabcolsep}{2pt}
  \begin{tabular}{@{}lcccc@{}}
    \toprule
    \textbf{Backbone} & \multicolumn{2}{c}{\(\boldsymbol{\Delta G}\)} & \multicolumn{2}{c}{\textbf{Probe}} \\
    \cmidrule(lr){2-3}\cmidrule(lr){4-5}
    & \shortstack{\textbf{CIFAR10}\\\textbf{Sym50}}
    & \shortstack{\textbf{CIFAR-10N}\\\textbf{Agg}}
    & \shortstack{\textbf{CIFAR10}\\\textbf{Sym50}}
    & \shortstack{\textbf{CIFAR-10N}\\\textbf{Agg}} \\
    \midrule
    ResNet-34 & 0.523 & 0.042 & 68.29 & 74.53 \\
    ResNet-50 & 0.945 & -0.034 & 68.77 & 74.93 \\
    MobileNetV2 & 1.066 & 0.269 & 64.55 & 70.86 \\
    \bottomrule
  \end{tabular}
  \caption{Additional backbone results for the preliminary analyses, reported outside the three representative architectures shown in \cref{fig:preliminary_summary}. \(\Delta G\) denotes the depth-sensitivity gap \(G^{(4)} - G^{(1)}\), and Probe denotes the layer1--4 mean linear-probe accuracy. Prototype-flipping metrics are reported once in the mechanism tables below.}
  \label{tab:appendix_extra_backbones}
\end{table}

\begin{table*}[htbp]
  \centering
  \footnotesize
  \setlength{\tabcolsep}{4pt}
  \begin{tabular}{llrrrr}
    \toprule
    \textbf{Backbone} & \textbf{Setting} & \(\boldsymbol{\Delta P_9}\) & \textbf{Flip4} & \textbf{Gap4} & \(\boldsymbol{\Delta}\)\textbf{Repair} \\
    \midrule
    ResNet-34 & CIFAR10-Sym50 & 0.049 & 0.998 & 3.841 & 0.41 \\
    ResNet-34 & CIFAR-10N-Agg & 0.005 & 0.999 & 5.559 & 0.99 \\
    ResNet-34 & CIFAR-10N-Worst & 0.058 & 0.998 & 3.931 & -3.00 \\
    \midrule
    ResNet-50 & CIFAR10-Sym50 & 0.001 & 0.997 & 3.363 & 0.71 \\
    ResNet-50 & CIFAR-10N-Agg & 0.006 & 1.000 & 5.009 & 2.30 \\
    ResNet-50 & CIFAR-10N-Worst & 0.026 & 0.997 & 3.550 & -0.01 \\
    \midrule
    WRN-28-10 & CIFAR10-Sym50 & 0.045 & 1.000 & 3.917 & 1.32 \\
    WRN-28-10 & CIFAR-10N-Agg & 0.037 & 1.000 & 5.124 & 3.13 \\
    WRN-28-10 & CIFAR-10N-Worst & 0.031 & 1.000 & 3.879 & 4.46 \\
    \midrule
    MobileNetV2 & CIFAR10-Sym50 & 0.080 & 0.209 & -0.738 & 3.56 \\
    MobileNetV2 & CIFAR-10N-Agg & 0.063 & 0.584 & 0.360 & 1.49 \\
    MobileNetV2 & CIFAR-10N-Worst & 0.077 & 0.396 & -0.297 & -0.24 \\
    \midrule
    DenseNet-121 & CIFAR10-Sym50 & -0.013 & 1.000 & 2.846 & -0.54 \\
    DenseNet-121 & CIFAR-10N-Agg & -0.110 & 1.000 & 4.052 & 10.99 \\
    \bottomrule
  \end{tabular}
  \caption{CIFAR-10 mechanism diagnostics for the additional backbones. \(\Delta P_9\) is the layer4--layer1 difference in the noisy-gradient projection ratio on mislabeled samples; Flip4 and Gap4 are the layer4 prototype-flip rate and attraction gap from P10; \(\Delta\)Repair is suffix34h recovery minus prefix12 recovery in P11. Positive \(\Delta\)Repair means late-stage surgery is less harmful than early-stage surgery.}
  \label{tab:appendix_mechanism_cifar10_extra}
\end{table*}

\textbf{Compute resources.}
All experiments were run on a local workstation with 8 NVIDIA GeForce RTX 4090 GPUs, each with 24GB memory.

\textbf{Existing assets.}
We use public benchmark datasets and previously published noisy-label baselines, and cite their original sources in the main paper. We follow the standard access and usage terms provided by the corresponding dataset and code maintainers.

\section{Reliability Analysis and Motivation}

This section provides the supplementary analysis behind TRACE's decoupled reliability design, including the theoretical view, extended preliminary evidence, and representation-level diagnostics.

\subsection{Reliability View of TRACE}
\label{app:reliability_view}

This subsection gives a simple theoretical view of why TRACE separates observed-label reliability from pseudo-target reliability and why shallow relations are useful as reliability anchors. The goal is not to prove a generalization guarantee for deep networks under arbitrary noise, but to make explicit the assumptions hidden by a single cleanliness coefficient.

\noindent\textbf{From label refurbishment to model structure.}
Label refurbishment first exposes a branch-coupling problem: the same sample-wise interpolation coefficient reduces observed-label weight and increases pseudo-target weight. This is reasonable only if low observed-label trust is evidence for high pseudo-target trust. However, the pseudo target is not an external supervision source; it is produced from predictions, teachers, or temporal ensembles built on the model's own representations. When noisy labels have already redirected these representations, observed-label unreliability and pseudo-target unreliability can coincide. This is why the main analysis moves from the refurbishment rule to model structure: it asks how noisy supervision changes representations across depth, and whether any part of the network remains stable enough to diagnose reliability. The observed shallow--deep contrast then motivates TRACE's two decisions: estimate observed-label reliability with a shallow structural anchor, and filter pseudo-target reliability with a separate confidence signal.

\noindent\textbf{Dual reliability instead of one interpolation score.}
Consider one training sample \(x_i\) with clean label \(y_i\), observed noisy label \(\hat{y}_i\), and pseudo-target class \(y_i^{\mathrm{pseudo}}=\arg\max_c q_{i,c}\). Let
\[
r_i^{\mathrm{obs}} =
P(\hat{y}_i = y_i \mid x_i),
\qquad
r_i^{\mathrm{pseudo}} =
P(y_i^{\mathrm{pseudo}} = y_i \mid x_i)
\]
denote the conditional reliability of the two supervision sources. A refurbishment target combines the observed-label branch and pseudo-target branch with nonnegative effective weights \(a_i\) and \(b_i\). Ignoring class-dependent loss curvature and focusing on whether each branch points to the clean class, the expected amount of correct supervision is
\[
S_i(a_i,b_i)=a_i r_i^{\mathrm{obs}} + b_i r_i^{\mathrm{pseudo}},
\]
whereas the expected amount of incorrect supervision is
\[
N_i(a_i,b_i)=a_i(1-r_i^{\mathrm{obs}})
+ b_i(1-r_i^{\mathrm{pseudo}}).
\]
The clean signal-to-noise ratio of the mixed target can therefore be summarized as
\begin{equation}
\mathrm{SNR}_i(a_i,b_i)
=
\frac{a_i r_i^{\mathrm{obs}} + b_i r_i^{\mathrm{pseudo}}}
{a_i(1-r_i^{\mathrm{obs}})+b_i(1-r_i^{\mathrm{pseudo}})+\varepsilon}.
\label{eq:app_snr}
\end{equation}
For fixed observed-label weight \(a_i\), increasing the pseudo-target weight in the idealized ratio is beneficial only when the pseudo target is more reliable than the observed-label branch. The stabilizer \(\varepsilon\) is only a numerical device; omitting it for the analytic comparison gives
\begin{equation}
\frac{\partial \mathrm{SNR}_i}{\partial b_i}
=
\frac{a_i\big(r_i^{\mathrm{pseudo}}-r_i^{\mathrm{obs}}\big)}
{\big[a_i(1-r_i^{\mathrm{obs}})+b_i(1-r_i^{\mathrm{pseudo}})\big]^2}.
\label{eq:app_snr_derivative}
\end{equation}
With the stabilizer retained, the numerator additionally contains the small numerical term \(\varepsilon r_i^{\mathrm{pseudo}}\), which should not be interpreted as statistical evidence for the pseudo branch. Thus, when \(a_i>0\), lowering trust in the observed label does not by itself justify increasing pseudo supervision; the pseudo branch improves the idealized mixture only if its own reliability is sufficiently high relative to the observed-label branch. This observation formalizes the failure mode discussed in the main text: a single cleanliness coefficient estimates \(r_i^{\mathrm{obs}}\), but then also controls \(b_i\), implicitly treating \(1-r_i^{\mathrm{obs}}\) as evidence for pseudo-target reliability.

The common interpolation rule \(a_i=\lambda_i\), \(b_i=1-\lambda_i\), with \(\lambda_i\) driven by a loss-based clean posterior, has exactly this coupling. If \(\lambda_i\) is small because the observed label looks unreliable, the pseudo branch becomes large even when \(r_i^{\mathrm{pseudo}}\) is low. In contrast, TRACE uses
\[
a_i=s_i^{\mathrm{obs}},
\qquad
b_i=(1-s_i^{\mathrm{obs}})s_i^{\mathrm{pseudo}},
\]
so the pseudo branch opens only when two conditions hold: observed-label reliability is low enough to need an alternative, and pseudo-target reliability is high enough to provide one. The total supervision strength
\[
w_i^{(0)}=a_i+b_i
=s_i^{\mathrm{obs}}+(1-s_i^{\mathrm{obs}})s_i^{\mathrm{pseudo}}
\]
also decreases when both branches are unreliable. The implemented loss uses \(w_i=\max(w_i^{(0)},w_{\min})\), as in \cref{eq:supervision_weighted_loss}, only to impose a small optimization lower bound. This is important because the right response to an unreliable observed label and an unreliable pseudo target is not to force a hard replacement target, but to reduce the sample's effective contribution. Under the view in \cref{eq:app_snr}, TRACE is therefore a branch-wise reliability filter rather than merely a different interpolation schedule.

This analysis is intentionally local and model-agnostic. It does not require the observed label and pseudo target to be independent: if both branches share a bias, then \(r_i^{\mathrm{pseudo}}\) can be low exactly when \(r_i^{\mathrm{obs}}\) is low, making the single-score coupling even less justified. It also matches the noisy-label literature's broader lesson that loss correction or sample selection is helpful only when the selected supervision remains aligned with the clean target \citep{DBLP:conf/nips/NatarajanDRT13,DBLP:conf/cvpr/PatriniRMNQ17,Han2018c}. TRACE keeps the useful small-loss prior for the observed-label branch, but prevents that prior from automatically becoming a pseudo-target trust score.

\subsection{Why Shallow Relations Can Serve as Anchors}
TRACE's observed-label reliability further asks whether a sample preserves its local relation pattern as representations become deeper. This design is motivated by the preliminary evidence that noisy supervision perturbs deeper layers more strongly while shallower relations remain comparatively stable. Here we give a small perturbation argument that connects this empirical pattern to the structural confidence in \cref{eq:structure_confidence}.

Let \(u_i^{(l)}\) be the unit-normalized clean feature of sample \(i\) at layer \(l\), and let the noisy-trained feature be
\[
\tilde{u}_i^{(l)}
=
\frac{u_i^{(l)}+\eta_i^{(l)}}{\|u_i^{(l)}+\eta_i^{(l)}\|_2},
\qquad
\|\eta_i^{(l)}\|_2\le \epsilon_l.
\]
For two samples \(i,j\), the cosine relation is \(A_{ij}^{(l)}=\langle u_i^{(l)},u_j^{(l)}\rangle\). Since the inner product of unit vectors is Lipschitz in each argument,
\begin{equation}
\left|
\left\langle \tilde{u}_i^{(l)},\tilde{u}_j^{(l)}\right\rangle
-\left\langle u_i^{(l)},u_j^{(l)}\right\rangle
\right|
\le
\left\|\tilde{u}_i^{(l)}-u_i^{(l)}\right\|_2
+
\left\|\tilde{u}_j^{(l)}-u_j^{(l)}\right\|_2.
\label{eq:app_cosine_lipschitz}
\end{equation}
When \(\epsilon_l<1\), normalization gives
\[
\left\|\tilde{u}_i^{(l)}-u_i^{(l)}\right\|_2
\le
\frac{2\epsilon_l}{1-\epsilon_l},
\]
and therefore
\begin{equation}
\left|
\tilde{A}_{ij}^{(l)}-A_{ij}^{(l)}
\right|
\le
\frac{4\epsilon_l}{1-\epsilon_l}.
\label{eq:app_relation_perturb}
\end{equation}
Thus, if shallow features have smaller perturbation \(\epsilon_l\), their pairwise relations are also more stable. For the sparse local relation matrix \(R^{(l)}=\mathcal{S}_k(A^{(l)})\), this statement requires a local margin condition because top-\(k\) off-diagonal selection is discontinuous at neighbor-ordering ties. Let \(\tau_l=4\epsilon_l/(1-\epsilon_l)\), and let \(m_i^{(l)}\) be the gap between the \(k\)-th retained off-diagonal affinity of sample \(i\) and the largest unretained affinity. If \(m_i^{(l)}>2\tau_l\), the top-\(k\) off-diagonal neighborhood of sample \(i\) is unchanged under the perturbation bound above; in this region, sparsification and symmetrization are fixed linear operations, so the row-wise sparse relation perturbation is controlled by the dense relation perturbation up to constants. When this margin condition fails, we do not claim a deterministic continuity bound; such samples correspond to locally ambiguous neighborhoods, for which relation drift is a useful warning signal rather than a certified perturbation measure.

Now decompose the measured cross-layer relation drift as
\begin{align}
\left\|\tilde{R}_i^{(l+1)}-\tilde{R}_i^{(l)}\right\|_2
&\le \left\|R_i^{(l+1)}-R_i^{(l)}\right\|_2 \notag\\
&\quad + \left\|\tilde{R}_i^{(l+1)}-R_i^{(l+1)}\right\|_2 \notag\\
&\quad
+ \left\|\tilde{R}_i^{(l)}-R_i^{(l)}\right\|_2 .
\label{eq:app_relation_drift_decomp}
\end{align}
The first term is the clean model's natural relation evolution across depth; the last two terms are noise-induced relation perturbations. This inequality should be read as a diagnostic decomposition rather than a lower bound: by itself it does not prove that the measured drift must increase whenever the deeper-layer perturbation increases. Under the additional empirical regime where the clean relation pattern changes smoothly and the perturbation terms do not cancel the deeper noisy redirection, larger deeper-layer perturbations explain larger measured shallow-to-deep drift. The main paper's gradient sensitivity and prototype-flipping analyses provide evidence that TRACE operates in this regime.

The structural confidence
\[
c_i^{\mathrm{str}}
=
\operatorname{Norm}\!\left(\exp(-\gamma \tilde{\delta}_i)\right),
\qquad
\delta_i=\sum_{l=1}^{L-1}
\frac{\left\|\tilde{R}_i^{(l+1)}-\tilde{R}_i^{(l)}\right\|_2}{\sqrt{B}},
\]
can therefore be interpreted as a monotone proxy for whether sample \(i\)'s local geometry remains stable as supervision propagates through the network. A high value does not mean the shallow feature is perfectly clean, nor does it turn shallow neighbors into labels. It only says that the sample has not undergone strong shallow-to-deep relational drift. This conservative use of shallow structure is consistent with structure-aware noisy-label methods \citep{Iscen2022LearningWN,Yi2022OnLC,DBLP:conf/aistats/WongsoGM23,DBLP:conf/icml/MainiMSLKZ23}, while differing from nearest-neighbor correction because TRACE uses the anchor to estimate reliability rather than to rewrite the pseudo target.

Combining the two arguments, TRACE can be read as a reliability-preserving control rule. The SNR view explains why the two supervision branches need separate scores, and the relation-stability view explains why shallow anchored drift is a useful signal for refining observed-label trust. Together they support the main design choice: low observed-label reliability should create an opportunity for pseudo supervision, but only a reliable pseudo target should be allowed to fill it.

\subsection{Hard-but-Clean Samples and the Scope of Relation Drift}
\label{app:hard_clean_scope}

Hard-but-clean samples clarify the intended scope of relation drift. A clean observed label does not imply that the corresponding sample must preserve an invariant neighborhood across depth. Samples near class boundaries, samples sharing visual attributes with other classes, and atypical instances within a class can undergo natural relational reorganization as the network builds increasingly task-specific features. Such drift may reflect sample difficulty rather than corruption. Consequently, \(c_i^{\mathrm{str}}\) in \cref{eq:structure_confidence} should be interpreted as a soft risk signal about the stability of the current representation path, not as a standalone clean/noisy decision.

TRACE incorporates this distinction directly into observed-label reliability. The structural confidence is blended with the loss posterior and modulated by prediction agreement, while \(\beta_t\) gradually activates the structure-aware terms. Thus, high relation drift alone is insufficient to determine \(s_i^{\mathrm{obs}}\): a hard-but-clean sample can still receive observed-label support through its loss-based evidence, and the agreement gate introduces no additional penalty when the two networks agree. Conversely, when the available signals jointly indicate uncertainty, the resulting reduction in \(s_i^{\mathrm{obs}}\) expresses lower confidence in the observed label rather than a declaration that it is incorrect.

Lower observed-label reliability also does not force a correction. By \cref{eq:weights}, the pseudo-branch weight is \(b_i=(1-s_i^{\mathrm{obs}})s_i^{\mathrm{pseudo}}\), so relation drift can create a need for alternative supervision but cannot make the pseudo target influential without separate pseudo-target evidence. If both the observed-label and pseudo-target signals are weak, \cref{eq:supervision_weighted_loss} reduces the sample's relative supervision strength subject to \(w_{\min}\), rather than committing to either source. TRACE therefore treats hard-but-clean ambiguity through continuous evidence fusion, independent pseudo-target filtering, and soft attenuation; relation drift modifies the level of trust assigned to a sample without being equated with label corruption.

\subsection{Extended Preliminary Analysis}

This subsection provides the extended version of the preliminary analysis summarized in the main paper, including additional backbone results and representation-level follow-up metrics under both synthetic and real-noise settings.

\subsubsection{Details for Figure~\ref{fig:preliminary_summary}}
\label{app:fig2_details}

Figure~\ref{fig:preliminary_summary} summarizes the preliminary representation diagnostics on CIFAR-10 with 20\%, 50\%, and 80\% symmetric noise. The three panels use the same representative backbones: ResNet-18, DenseNet-121, and WRN-28-10. The purpose is to separate three related effects of noisy supervision: how strongly gradients are perturbed, whether mislabeled samples move toward noisy prototypes, and whether the resulting features remain linearly useful.

\textbf{Panel (i): layerwise gradient sensitivity.}
This panel reports the gradient-sensitivity ratio \(G^{(l)}\) defined in the main text. Larger values mean that the layer's cross-entropy gradient under noisy supervision deviates more from the clean-supervision gradient. The key comparison is across depth and noise rate: deeper layers show larger increases as the corruption rate grows, indicating stronger noisy redirection in late representations.

\textbf{Panel (ii): layerwise prototype flipping.}
This panel measures the fraction of mislabeled samples that are closer to their observed-label prototype than to their clean-label prototype. The contrast between layer1 and layer4 shows whether noisy labels merely perturb features or actively pull deep representations toward incorrect class anchors. The near-saturated layer4 flip rates for the representative backbones indicate that deep features can become strongly organized around noisy labels.

\textbf{Panel (iii): linear-probe degradation.}
This panel reports the layer1--4 mean linear-probe accuracy after freezing the learned representations. Lower probe accuracy at higher noise rates means that the representation has become less directly aligned with clean semantics. Together with panels (i) and (ii), this supports the use of shallow relations as anchors and motivates separating observed-label reliability from pseudo-target reliability.

\subsubsection{Details for Figure~\ref{fig:drift_comparison}}
\label{app:fig_drift_details}

Figure~\ref{fig:drift_comparison} is a conditional reliability diagnostic on CIFAR-100 with 50\% symmetric label noise. Clean labels are used only to evaluate whether the observed label is correct and are not available to the training procedure. For every training sample, we obtain the loss posterior \(c_i^{\mathrm{loss}}\) and the shallow-to-deep relation drift \(\delta_i\) from the same model-scoring pass. The displayed loss-confidence strata are \(\{0\}\), \((0,0.2)\), \([0.2,0.5)\), \([0.5,0.8)\), \([0.8,0.98)\), and \([0.98,1]\).

\textbf{Within-stratum drift comparison.}
Within each \(c_i^{\mathrm{loss}}\) stratum, samples are ranked by \(\delta_i\) and split at the stratum-wise median. The lower and upper halves are denoted low drift and high drift, respectively; because \(c_i^{\mathrm{str}}\) is a monotone decreasing transformation of drift before dual-network fusion, these groups correspond to relatively high and low structural confidence. Each bar reports
\[
\frac{1}{|\mathcal{G}|}\sum_{i\in\mathcal{G}}
\mathbf{1}\!\left[y_i^{\mathrm{obs}}=y_i^\star\right],
\]
where \(\mathcal{G}\) is the corresponding loss-stratum/drift group and \(y_i^\star\) is the clean reference label used only for analysis.

\textbf{Interpretation and scope.}
The comparison controls loss confidence at the resolution of the displayed strata rather than claiming exact sample matching. Its purpose is to test whether relation drift provides conditional information beyond \(c_i^{\mathrm{loss}}\). The separation is strongest where loss-based evidence is weak and naturally contracts in high-confidence strata because observed-label correctness approaches its ceiling. Accordingly, Figure~\ref{fig:drift_comparison} supports using \(c_i^{\mathrm{str}}\) as a complementary modifier of the loss posterior, not as a standalone label-correctness certificate.


\subsubsection{Additional Backbone Results Beyond Figure~\ref{fig:preliminary_summary}}
\label{app:extra_backbones}

\cref{tab:appendix_extra_backbones} clarifies why the main text focuses on ResNet-18, DenseNet-121, and WRN-28-10. ResNet-34 and ResNet-50 broadly follow the same trends as the main-text backbones, although the CIFAR-10N-Agg depth gap is weak for ResNet-34 and slightly negative for ResNet-50. MobileNetV2 provides a more substantial deviation in the prototype-flipping analysis, which is reported in \cref{tab:appendix_mechanism_cifar10_extra}: its layer4 flip rate is only 0.209 on CIFAR-10 with 50\% symmetric noise and 0.584 on CIFAR-10N-Agg, far below the near-saturated behavior observed for the three representative backbones in the main paper. We therefore use the main text to emphasize the stable pattern shared by the three selected architectures, while retaining the remaining architectures here to document the broader empirical boundary.

For compact table captions, we use three protocol labels for auxiliary mechanism diagnostics: P9 denotes the noisy-gradient projection ratio on mislabeled samples, P10 denotes prototype-attraction diagnostics, and P11 denotes checkpoint-surgery repair.

\begin{table*}[htbp]
  \centering
  \footnotesize
  \setlength{\tabcolsep}{4pt}
  \begin{tabular}{llrrrr}
    \toprule
    \textbf{Backbone} & \textbf{Setting} & \(\boldsymbol{\Delta P_9}\) & \textbf{Flip4} & \textbf{Gap4} & \(\boldsymbol{\Delta}\)\textbf{Repair} \\
    \midrule
    ResNet-34 & CIFAR100-Sym50 & -0.014 & 0.999 & 3.546 & -0.25 \\
    ResNet-50 & CIFAR100-Sym50 & -0.019 & 0.999 & 4.195 & 0.19 \\
    \midrule
    WRN-28-10 & CIFAR100-Sym50 & 0.064 & 1.000 & 7.098 & 0.10 \\
    WRN-28-10 & CIFAR-100N & 0.123 & 1.000 & 7.255 & 0.46 \\
    \midrule
    MobileNetV2 & CIFAR100-Sym50 & 0.013 & 0.826 & 0.789 & 0.53 \\
    MobileNetV2 & CIFAR-100N & 0.001 & 0.884 & 1.243 & 0.19 \\
    \midrule
    DenseNet-121 & CIFAR100-Sym50 & -0.017 & 0.999 & 3.396 & 0.79 \\
    DenseNet-121 & CIFAR-100N & -0.025 & 0.999 & 4.832 & 0.87 \\
    \bottomrule
  \end{tabular}
  \caption{CIFAR-100 mechanism diagnostics for the additional backbones, using the same metrics as \cref{tab:appendix_mechanism_cifar10_extra}. These results extend the preliminary mechanism checks beyond the CIFAR-10 settings already summarized in \cref{tab:appendix_extra_backbones}.}
  \label{tab:appendix_mechanism_cifar100_extra}
\end{table*}

The additional mechanism tables are meant to define the empirical boundary rather than to strengthen the claim into a universal depth law. P9 generally supports deeper-layer gradient redirection for ResNet/WRN/MobileNet-style backbones, but DenseNet-121 is often weaker or negative in the layer4--layer1 contrast. P10 is the most stable mechanism signal for ResNet, WRN, and DenseNet, whereas MobileNetV2 is a clear CIFAR-10 exception and becomes only moderately supportive on CIFAR-100. P11 remains an auxiliary intervention check because checkpoint surgery is harsh; its sign is useful mainly as relative evidence for whether deeper-stage repair is preferable to early-stage repair.

\subsection{Additional Representation Analysis}
\label{app:representation_followup}

\begin{table}[htbp]
  \centering
  \footnotesize
  \setlength{\tabcolsep}{6pt}
  \begin{tabular}{lcc}
    \toprule
    \textbf{Setting} & \textbf{CKA} & \textbf{FDR} \\
    \midrule
    CIFAR10-Sym20 & 0.7223 & 0.5780 \\
    CIFAR10-Sym50 & 0.6544 & 0.4339 \\
    CIFAR10-Sym80 & 0.5275 & 0.2877 \\
    CIFAR-10N-Agg & 0.7606 & 0.7186 \\
    CIFAR-10N-Worst & 0.6805 & 0.4372 \\
    CIFAR100-Sym20 & 0.6975 & 0.3849 \\
    CIFAR100-Sym50 & 0.6340 & 0.3305 \\
    CIFAR100-Sym80 & 0.5026 & 0.2828 \\
    CIFAR-100N & 0.7186 & 0.4186 \\
    \bottomrule
  \end{tabular}
  \caption{Representation-level follow-up metrics on ResNet-18. We report layer1--4 mean values for clean--noisy representation similarity and class separability. Both metrics deteriorate with increasing synthetic noise and under harder real-noise conditions.}
  \label{tab:repr_followup_metrics}
\end{table}

\begin{table}[htbp]
  \centering
  \footnotesize
  \setlength{\tabcolsep}{4pt}
  \begin{tabular}{llcc}
    \toprule
    \textbf{Backbone} & \textbf{Setting} & \textbf{CKA} & \textbf{FDR} \\
    \midrule
    ResNet-34 & CIFAR10-Sym50 & 0.6125 & 0.4029 \\
    ResNet-34 & CIFAR-10N-Agg & 0.7786 & 0.8044 \\
    ResNet-34 & CIFAR-10N-Worst & 0.6675 & 0.4304 \\
    ResNet-34 & CIFAR100-Sym50 & 0.5552 & 0.2969 \\
    \midrule
    ResNet-50 & CIFAR10-Sym50 & 0.7000 & 0.4127 \\
    ResNet-50 & CIFAR-10N-Agg & 0.7783 & 0.6502 \\
    ResNet-50 & CIFAR-10N-Worst & 0.7402 & 0.4342 \\
    \midrule
    WRN-28-10 & CIFAR10-Sym50 & 0.6397 & 0.5356 \\
    WRN-28-10 & CIFAR-10N-Agg & 0.7904 & 1.3144 \\
    WRN-28-10 & CIFAR-10N-Worst & 0.6590 & 0.5685 \\
    WRN-28-10 & CIFAR100-Sym50 & 0.5839 & 0.4041 \\
    WRN-28-10 & CIFAR-100N & 0.7019 & 0.5870 \\
    \midrule
    MobileNetV2 & CIFAR10-Sym50 & 0.6702 & 0.7133 \\
    MobileNetV2 & CIFAR-10N-Agg & 0.7567 & 0.8522 \\
    MobileNetV2 & CIFAR-10N-Worst & 0.6990 & 0.7032 \\
    MobileNetV2 & CIFAR100-Sym50 & 0.6460 & 0.4252 \\
    MobileNetV2 & CIFAR-100N & 0.7228 & 0.4902 \\
    \midrule
    DenseNet-121 & CIFAR10-Sym50 & 0.6368 & 0.3335 \\
    DenseNet-121 & CIFAR-10N-Agg & 0.8299 & 0.6660 \\
    DenseNet-121 & CIFAR100-Sym50 & 0.6392 & 0.3288 \\
    DenseNet-121 & CIFAR-100N & 0.7499 & 0.4327 \\
    \bottomrule
  \end{tabular}
  \caption{Additional representation-level metrics for non-ResNet-18 backbones. To avoid duplicating \cref{tab:repr_followup_metrics}, this table reports only the extra architectures and focuses on clean--noisy CKA and FDR over layer1--4.}
  \label{tab:appendix_extra_repr_metrics}
\end{table}

\textbf{Class separability degradation.}
We ask whether the depth-dependent degradation diagnosed is also visible through class separability. For layer feature \(T_l\), we use the Fisher discriminant ratio
\begin{equation}
\mathrm{FDR}_l = \frac{\mathrm{tr}(S_b)}{\mathrm{tr}(S_w) + \varepsilon},
\end{equation}
where \(S_b\) and \(S_w\) are the between-class and within-class scatter matrices. Larger values indicate that classes remain well separated. As summarized in \cref{tab:repr_followup_metrics}, the layer1--4 mean FDR on ResNet-18 drops from 0.5780 to 0.4339 to 0.2877 as CIFAR-10 symmetric noise increases from 20\% to 80\%, and from 0.3849 to 0.3305 to 0.2828 on CIFAR-100. Real-noise results show the same trend: CIFAR-10N-Worst is substantially below CIFAR-10N-Agg (0.4372 versus 0.7186). \cref{tab:appendix_extra_repr_metrics} reports the same CKA/FDR diagnostics for the remaining backbones wherever the corresponding representation suite is available.

\begin{table}[htbp]
  \centering
  \footnotesize
  \setlength{\tabcolsep}{4pt}
  \begin{tabular}{lcccc}
    \toprule
    \textbf{Setting} & \textbf{Layer1} & \textbf{Layer4} & \textbf{Prefix12} & \textbf{Suffix34h} \\
    \midrule
    CIFAR10-Sym50 & -48.80 & -44.61 & -49.01 & -46.14 \\
    CIFAR-10N-Rand1 & -71.84 & -67.63 & -73.86 & -62.08 \\
    CIFAR100-Sym50 & -39.86 & -39.53 & -40.01 & -39.69 \\
    CIFAR-100N & -51.85 & -51.94 & -52.47 & -51.54 \\
    \bottomrule
  \end{tabular}
  \caption{Checkpoint-surgery recovery relative to the noisy baseline. Prefix12 transplants clean layer1 and layer2 into the noisy model; Suffix34h transplants clean layer3, layer4, and the classifier head. Although the intervention is harsh and absolute values remain negative, deeper-stage repair is usually less harmful than early-stage repair.}
  \label{tab:repair_followup}
\end{table}

\textbf{Layerwise repair preference.}
Starting from a noisy checkpoint, we transplant selected stages from the corresponding clean model and measure recovery relative to the noisy baseline:
\begin{equation}
\mathrm{recover}(v) =
\mathrm{Acc}(\text{repair variant } v) - \mathrm{Acc}(\text{noisy base}).
\end{equation}
The checkpoint-surgery protocol is intentionally simple and therefore harsh, so the absolute numbers in \cref{tab:repair_followup} should be interpreted only directionally. The variants named in the table replace the corresponding clean stages into the noisy checkpoint: Layer1 and Layer4 replace a single residual stage, Prefix12 replaces layer1--2, and Suffix34h replaces layer3--4 plus the classifier head. Even so, the relative ordering is informative. On CIFAR-10 with 50\% symmetric noise, repairing layer4 is less harmful than repairing layer1 (-44.61 versus -48.80), and repairing Suffix34h is better than repairing Prefix12 (-46.14 versus -49.01). The same ordering becomes clearer on CIFAR-10N-Rand1, where Suffix34h repair outperforms Prefix12 repair by more than 11 points.

\textbf{Clean--noisy representation mismatch.}
Finally, we compare noisy representations with a clean reference model through linear CKA:
\begin{equation}
\mathrm{CKA}(X, Y) =
\frac{\|X_c^{\top} Y_c\|_F^2}
{\sqrt{\|X_c^{\top} X_c\|_F^2 \cdot \|Y_c^{\top} Y_c\|_F^2}},
\end{equation}
where \(X_c\) and \(Y_c\) are centered activations from the clean and noisy models. Lower CKA indicates that noisy representations have drifted farther from their clean counterparts. \cref{tab:repr_followup_metrics} shows the expected progression: on CIFAR-10 symmetric noise, the layer1--4 mean CKA declines from 0.7223 to 0.6544 to 0.5275 as the noise rate increases; on CIFAR-100, it declines from 0.6975 to 0.6340 to 0.5026. Real noise again tells the same story, with CIFAR-10N-Agg remaining closer to the clean reference than CIFAR-10N-Worst (0.7606 versus 0.6805).

\section{Experimental Protocols}

\subsection{Additional Experimental Details}

The main evaluation metric is classification accuracy. Unless otherwise stated, CIFAR-style results report the average test accuracy over the last 10 epochs. In ablation and sensitivity tables with \textbf{Best} and \textbf{Last} columns, Best denotes the highest test accuracy reached by any checkpoint during training, while Last denotes the average test accuracy over the final 10 epochs. Large-scale real-noise experiments follow the reporting protocol of the corresponding base learner.

For synthetic noise, we use CIFAR-10/100 \citep{krizhevsky2009learning} with 20\%, 50\%, and 80\% symmetric noise \citep{Song2019}, 40\% pair/asymmetric noise \citep{Song2019}, and 40\% instance-dependent noise \citep{Chen2020BeyondCA}. For real-world noisy datasets, we use CIFAR-10N/CIFAR-100N \citep{DBLP:conf/iclr/WeiZ0L0022}, including CIFAR-10N-Agg, CIFAR-10N-Rand1--3, CIFAR-10N-Worst, and CIFAR-100N; we also evaluate WebVision \citep{DBLP:journals/corr/abs-1708-02862}, Food-101N \citep{lee2018cleannet}, and Clothing1M under the reporting protocols used by the corresponding base methods.

TRACE is attached to representative refurbishment-based learners without changing their pseudo-target generators or training pipelines. For the RoLR instantiation, we reuse the GMM loss-clean posterior, add shallow structural confidence and dual-network agreement for observed-label reliability, and apply a separate confidence score to the pseudo branch. Augmentations and reproduced plug-in variants follow the corresponding base protocols. Experiments vary only the dataset, noise setting, base learner, or explicitly named ablation factors unless otherwise specified.

\subsubsection{Component Ablation Variants}
\cref{tab:main_ablation} isolates the components of TRACE while keeping the RoLR pseudo-target generator, data pipeline, backbone, and training schedule unchanged. The \(+s_i^{\mathrm{obs}}\) variant replaces RoLR's original loss-based cleanliness coefficient with the refined observed-label reliability score, but does not introduce the separate pseudo-target reliability score or the supervision-strength weighting. TRACE-CE uses the full source-weighted target with \(s_i^{\mathrm{obs}}\) and \(s_i^{\mathrm{pseudo}}\), but optimizes the standard soft-target cross-entropy without the sample-wise weight \(w_i\) in \cref{eq:supervision_weighted_loss}. The TRACE-CE w/o \(c_i^{\mathrm{str}}\) variant removes the shallow structural confidence from \(s_i^{\mathrm{obs}}\), so observed-label reliability falls back to the loss posterior together with the agreement gate. The TRACE-CE w/o \(g_i^{\mathrm{agr}}\) variant disables the dual-network agreement gate by setting the agreement multiplier to one. The TRACE-CE w/o \(s_i^{\mathrm{pseudo}}\) variant removes the separate pseudo-target reliability assessment and uses the complementary pseudo-branch weight \(1-s_i^{\mathrm{obs}}\). The final TRACE row restores the supervision-strength weighting, so the ablation changes only the reliability interface and loss weighting rather than the underlying training pipeline.

\subsubsection{Synthetic Noise Construction}
For CIFAR-style synthetic-noise experiments, we first construct a fixed noisy-label file for each dataset, noise type, noise rate, and random seed, and all compared methods are trained with the same observed labels. Let \(y_i\in\{1,\ldots,C\}\) be the clean class of sample \(x_i\), and let \(\hat y_i\) be the label actually used for training.

\emph{Symmetric noise} corrupts labels independently of the image content and treats all wrong classes uniformly \citep{Jiang2017,Malach2017,Song2019}. At noise rate \(\xi\), a sample keeps its clean label with probability \(1-\xi\); otherwise its label is replaced by one of the other \(C-1\) classes uniformly at random. Equivalently, the class-transition matrix has diagonal probability \(1-\xi\) and off-diagonal probability \(\xi/(C-1)\). This setting models unstructured random annotation errors.

\emph{Asymmetric (pair) noise} is still class-conditional, but the wrong label is not chosen uniformly \citep{Jiang2017,Malach2017,Song2019}. Instead, each corrupted source class is mapped to a semantically related target class, so the transition matrix places most corrupted mass on a single paired class. For CIFAR-10, this follows the common benchmark mapping used by refurbishment baselines, such as bird \(\rightarrow\) airplane, cat \(\leftrightarrow\) dog, deer \(\rightarrow\) horse, and truck \(\rightarrow\) automobile. For CIFAR-100, we use the corresponding benchmark pair-noise labels from the same evaluation protocol. This noise type is harder than symmetric noise because the wrong labels are visually plausible rather than arbitrary.

\emph{Instance-dependent noise} allows the corruption distribution to depend on the input \(x_i\), following the standard synthetic protocol for instance-dependent label noise \citep{Chen2020BeyondCA}. Instead of using a single class-transition matrix shared by every sample in a class, each example receives its own transition vector \(T_i\). The clean class keeps the remaining probability mass, while candidate wrong classes receive probabilities determined from sample-dependent scores. We then sample \(\hat y_i\) from this per-example distribution. Thus, visually ambiguous or feature-specific samples can have different corruption tendencies even when they share the same clean class, which makes the setting closer to realistic annotation mistakes than class-conditional noise.

To test reliability directly, we also measure pseudo-target accuracy on all samples, low-clean samples, and low-clean$+$noisy samples. Low-clean samples are those assigned low observed-label reliability by the corresponding method, and low-clean$+$noisy samples further restrict this subset to examples whose observed label is actually corrupted. When ground-truth clean labels are available for analysis, pseudo-target accuracy is computed by comparing the selected pseudo target with the clean label. Follow-noisy is computed only on observed-noisy samples and measures how often the pseudo target agrees with the noisy observed label. HC Wrong denotes high-confidence pseudo targets that are nevertheless incorrect, so lower values indicate safer pseudo supervision. Representation-level follow-up metrics are reported to connect the final method back to the preliminary analysis.

The qualitative examples in \cref{fig:case_study_cifar10} are shown from CIFAR-10's native \(32\times32\) images; their limited visual sharpness reflects the dataset resolution rather than a rendering artifact.

\section{Additional Experimental Results}

This section collects supplementary experimental results beyond the main paper.

\subsection{Results under 20\% Pair and Instance-dependent Noise}

To complement the 40\% pair and instance-dependent noise results in \cref{tab:main_1}, we additionally evaluate RoLR and TRACE under the milder 20\% settings on CIFAR-10 and CIFAR-100. We keep the backbone, noisy-label realization, training schedule, and reporting protocol matched within each pair, changing only the reliability mechanism.

\begin{table}[htbp]
  \centering
  \footnotesize
  \setlength{\tabcolsep}{3.5pt}
  \begin{tabular}{lcccc}
    \toprule
    \multirow{2}{*}{\textbf{Method}} &
    \multicolumn{2}{c}{\textbf{CIFAR-10}} &
    \multicolumn{2}{c}{\textbf{CIFAR-100}} \\
    \cmidrule(lr){2-3}\cmidrule(lr){4-5}
    & \textbf{Pair-20\%} & \textbf{Ins-20\%}
    & \textbf{Pair-20\%} & \textbf{Ins-20\%} \\
    \midrule
    RoLR & 95.0 & 95.4 & 78.6 & 79.1 \\
    \quad + TRACE & \textbf{95.7} & \textbf{95.8} & \textbf{80.1} & \textbf{80.3} \\
    \bottomrule
  \end{tabular}
  \caption{Test accuracy (\%) of RoLR and TRACE under 20\% pair and instance-dependent noise.}
  \label{tab:appendix_pair_ins_20}
\end{table}

\subsection{Additional Hyperparameter Sensitivity}

\cref{tab:appendix_sensitivity_hyperparameters} reports the remaining one-factor-at-a-time sensitivity settings on CIFAR-100 with 50\% symmetric noise. These factors control the structural anchor, scheduling, agreement attenuation, and minimum supervision strength. They are kept in the appendix because they are implementation-facing checks rather than the two branch-defining hyperparameters emphasized in the main text.

\begin{table*}[htbp]
  \centering
  \footnotesize
  \setlength{\tabcolsep}{4pt}
  \begin{tabular}{lccccc}
    \toprule
    \textbf{Factor} & \textbf{Value} & \textbf{Best} & \textbf{Last} & \textbf{Pseudo Acc.} & \textbf{HC Wrong} \\
    \midrule
     & \(20\) & \underline{76.84} & \underline{76.10} & \underline{84.52} & \underline{6.24} \\
    Relation neighbors \(k\) & \(50\) & \textbf{76.95} & \textbf{76.22} & \textbf{84.70} & \textbf{6.16} \\
     & \(100\) & 76.76 & 76.03 & 84.35 & 6.31 \\
    \midrule
     & \(2.5\) & \underline{76.88} & \underline{76.12} & \underline{84.61} & \underline{6.22} \\
    Drift temperature \(\gamma\) & \(5.0\) & \textbf{76.95} & \textbf{76.22} & \textbf{84.70} & \textbf{6.16} \\
     & \(10.0\) & 76.71 & 75.96 & 84.29 & 6.38 \\
    \midrule
     & \(20\) & \underline{76.86} & \underline{76.11} & \underline{84.46} & 6.32 \\
    Pseudo-score start epoch & \(30\) & \textbf{76.95} & \textbf{76.22} & \textbf{84.70} & \underline{6.16} \\
     & \(40\) & 76.69 & 75.91 & 84.28 & \textbf{6.09} \\
    \midrule
     & \(0.25\) & 76.74 & 75.99 & 84.33 & 6.41 \\
    Agreement penalty \(\lambda_{\mathrm{dis}}\) & \(0.5\) & \textbf{76.95} & \textbf{76.22} & \textbf{84.70} & \underline{6.16} \\
     & \(0.75\) & \underline{76.82} & \underline{76.08} & \underline{84.55} & \textbf{6.04} \\
    \midrule
     & \(0.0\) & 76.63 & 75.87 & 84.26 & 6.47 \\
    Minimum sample weight \(w_{\min}\) & \(0.2\) & \textbf{76.95} & \textbf{76.22} & \textbf{84.70} & \underline{6.16} \\
     & \(0.4\) & \underline{76.79} & \underline{76.04} & \underline{84.49} & \textbf{6.02} \\
    \bottomrule
  \end{tabular}
  \caption{Additional sensitivity analysis on CIFAR-100 with 50\% symmetric noise. Each row varies one hyperparameter while keeping all other defaults fixed. HC Wrong denotes the high-confidence pseudo-target error rate, where lower is better.}
  \label{tab:appendix_sensitivity_hyperparameters}
\end{table*}

These additional sweeps support the default choices used in the main experiments. A moderate neighborhood size, temperature, pseudo-score start time, agreement penalty, and minimum sample weight provide the best or most balanced accuracy--reliability trade-off. More conservative pseudo filtering can slightly reduce high-confidence errors, but it also lowers best/last accuracy and pseudo-target accuracy, so we keep the middle settings as the default configuration. Across the main and appendix sweeps, performance changes remain small and smooth, indicating that TRACE's two-source reliability interface is not tied to a fragile hyperparameter setting.

\subsection{Additional Reliability Statistics}

\cref{tab:appendix_cross_dataset_reliability} extends the main-text reliability summary with three complementary views of pseudo-target quality. All Pseudo Acc. measures global pseudo-target accuracy, Low-clean Pseudo Acc. focuses on samples assigned low observed-label reliability by the corresponding method, and Low-clean+Noisy Pseudo Acc. further restricts that subset to samples whose observed label is actually corrupted. The last view most directly tests the supervision used when refurbishment needs a replacement target. Because the low-clean subset is selected separately by RoLR and TRACE, its two entries are diagnostic rather than a matched-sample comparison.

TRACE improves Low-clean+Noisy Pseudo Acc. in every reported setting. The gains are largest across the three CIFAR-100 symmetric-noise rates, where the larger label space makes reliable replacement harder, and remain positive under CIFAR-10 and human-annotation noise. Global pseudo-target accuracy also generally improves, although the small decreases on CIFAR100-Sym20 and CIFAR-100N mark the boundary of this effect. Together, the results show that TRACE's most consistent benefit occurs on actually corrupted samples that depend on the pseudo branch, rather than uniformly improving every method-dependent subset.

\begin{table*}[htbp]
  \centering
  \footnotesize
  \setlength{\tabcolsep}{5pt}
  \begin{tabular}{lccc}
    \toprule
    \textbf{Setting} & \textbf{All Pseudo Acc.} & \textbf{Low-clean Pseudo Acc.} & \textbf{Low-clean+Noisy Pseudo Acc.} \\
    & \textbf{RoLR / TRACE} & \textbf{RoLR / TRACE} & \textbf{RoLR / TRACE} \\
    \midrule
    \multicolumn{4}{l}{\textit{Synthetic Noise}} \\
    CIFAR10-Sym20 & 98.38 / 98.58 & 96.05 / 93.20 & 96.89 / 98.84 \\
    CIFAR10-Sym50 & 96.34 / 97.16 & 95.93 / 94.51 & 96.42 / 97.52 \\
    CIFAR10-Sym80 & 93.63 / 94.68 & 94.82 / 93.55 & 95.17 / 95.21 \\
    CIFAR100-Sym20 & 93.51 / 93.05 & 73.14 / 74.85 & 76.51 / 93.02 \\
    CIFAR100-Sym50 & 83.47 / 84.56 & 70.75 / 74.15 & 72.92 / 84.70 \\
    CIFAR100-Sym80 & 44.11 / 53.66 & 38.20 / 50.20 & 42.41 / 54.09 \\
    \midrule
    \multicolumn{4}{l}{\textit{Real Noise}} \\
    CIFAR-10N-Agg & 96.58 / 97.42 & 90.96 / 89.44 & 93.02 / 94.07 \\
    CIFAR-10N-Rand1 & 95.80 / 96.89 & 93.51 / 93.07 & 95.48 / 96.16 \\
    CIFAR-10N-Rand2 & 95.70 / 96.82 & 93.98 / 92.73 & 95.42 / 95.70 \\
    CIFAR-10N-Rand3 & 94.76 / 96.81 & 94.92 / 92.93 & 95.81 / 96.02 \\
    CIFAR-10N-Worst & 91.23 / 94.39 & 92.87 / 94.36 & 93.83 / 95.78 \\
    CIFAR-100N & 71.86 / 71.22 & 57.43 / 57.82 & 59.68 / 60.38 \\
    \bottomrule
  \end{tabular}
  \caption{Full cross-dataset pseudo-target accuracy under all evaluated noise conditions. The low-clean+noisy subset isolates the cases in which the observed label is actually corrupted and pseudo supervision matters most.}
  \label{tab:appendix_cross_dataset_reliability}
\end{table*}

\subsection{CIFAR-100N Hard-Regime Diagnostics}
\cref{tab:cifar100n_hard_regime} provides subset-level diagnostics for the hardest real-noise regime. CIFAR-100N is where the main-text reliability gain is positive but small, and the subset statistics show that pseudo targets can still follow noisy observed labels. Follow-noisy measures pseudo-target agreement with the noisy observed label, and is computed only where the observed label is noisy. On observed-noisy samples, pseudo-target accuracy drops to 30.93\%, while follow-noisy remains 51.72\%, reinforcing that pseudo-target reliability is related to sample cleanliness but cannot be reduced to it.

\begin{table}[htbp]
  \centering
  \footnotesize
  \setlength{\tabcolsep}{6pt}
  \begin{tabular}{lccc}
    \toprule
    \textbf{Subset} & \textbf{Count} & \textbf{Pseudo Acc.} & \textbf{Follow-noisy} \\
    \midrule
    All & 50000 & 71.22 & 51.72 \\
    Low-clean & 10466 & 57.82 & 4.91 \\
    Observed-noisy & 20100 & 30.93 & 51.72 \\
    Low-clean+noisy & 9993 & 60.38 & 4.91 \\
    \bottomrule
  \end{tabular}
  \caption{CIFAR-100N hard real-noise reliability under the proposed method.}
  \label{tab:cifar100n_hard_regime}
\end{table}

\begin{center}
  \begin{minipage}{\columnwidth}
    \centering
    \includegraphics[width=\columnwidth]{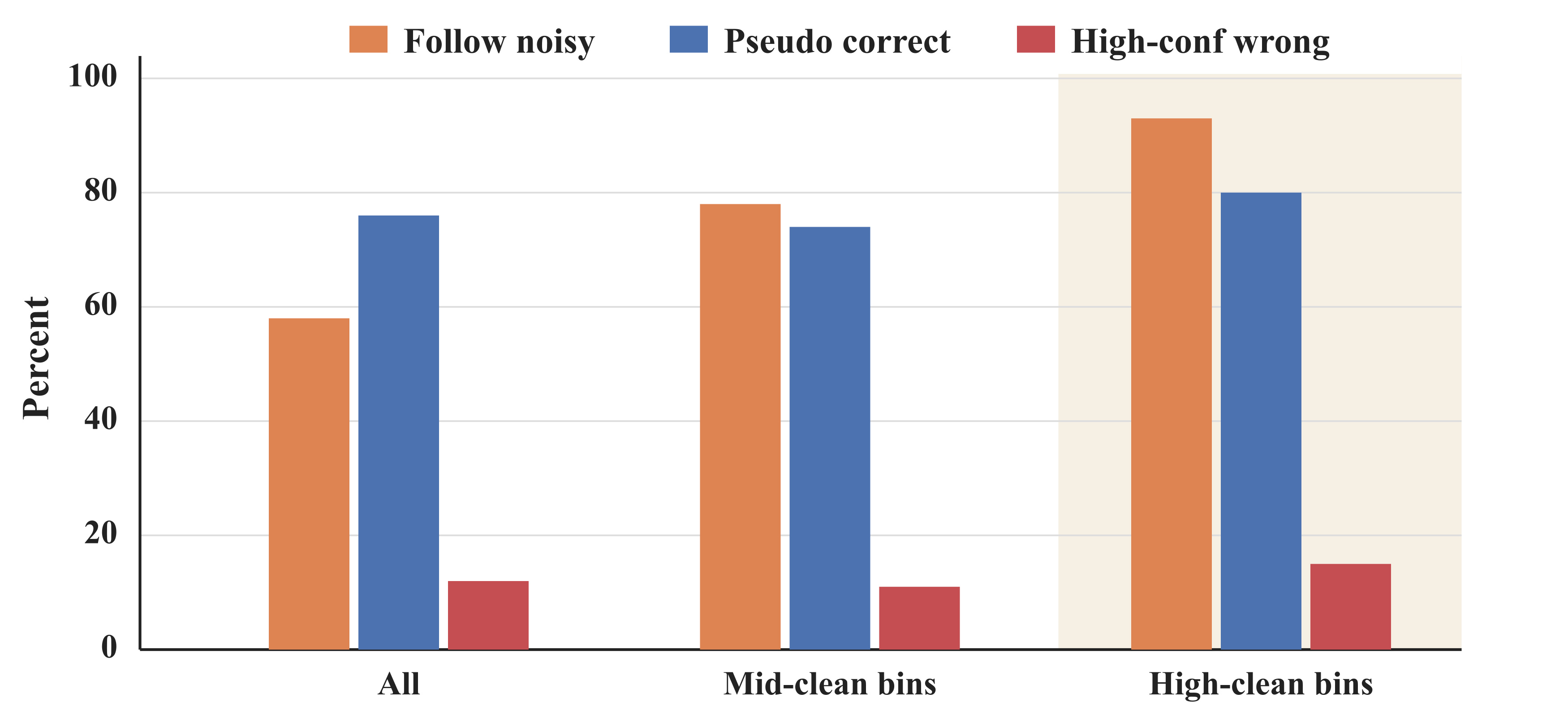}
    \captionof{figure}{Pseudo-target mismatch on a random 800-image subset of the Clothing1M noisy training split. GPT-5.6 API annotations are used as surrogate reference labels for this diagnostic. Higher estimated cleanliness does not eliminate noisy-label following or high-confidence pseudo-target errors.}
    \label{fig:clothing1m_pseudo_target_mismatch}
  \end{minipage}
\end{center}

\subsection{Clothing1M Cross-Dataset Diagnostic}
\label{app:clothing1m_diagnostic}

We further examine whether the pseudo-target mismatch in \cref{fig:intro_pseudo_target_unreliable} also appears under large-scale real-world label noise. We randomly sample 800 images from the noisy training split of Clothing1M. Its training images were collected from online shopping websites, and their noisy labels were derived from surrounding text and label keywords rather than image-level manual verification \cite{DBLP:conf/cvpr/XiaoXYHW15}. The benchmark also contains a clean subset formed by manually refining 72,409 image labels, whereas the 800 images analyzed here come from the million-image noisy split. We therefore query the GPT-5.6 API to assign one of the 14 Clothing1M categories to each sampled image and use these annotations as surrogate reference labels for this diagnostic. Following the analysis in \cref{fig:intro_pseudo_target_unreliable}, we report how often the pseudo target follows the noisy observed label, matches the surrogate reference label, or remains wrong at high confidence across estimated-cleanliness strata.

As shown in \cref{fig:clothing1m_pseudo_target_mismatch}, the high-clean bins exhibit the clearest mismatch: pseudo targets follow the noisy observed labels on 93\% of samples, but match the surrogate reference labels on only 80\%, while 15\% remain high-confidence errors. A similar gap appears in the mid-clean bins. These results qualitatively reproduce the phenomenon in \cref{fig:intro_pseudo_target_unreliable} under a distinct, real-world noise source: a high cleanliness estimate does not by itself certify the pseudo target. Because the reference labels are model-generated and the analysis covers only 800 sampled images, this experiment is intended as evidence of recurrence rather than a dataset-wide prevalence estimate or a substitute for human-verified ground truth.

\section{Figure Details}

\subsection{Details for Figure~\ref{fig:intro_pseudo_target_unreliable}}
\label{app:fig1_details}

Figure~\ref{fig:intro_pseudo_target_unreliable} is constructed from the CIFAR-100N real-noise setting under the noisy condition. The statistics are exported from the same pseudo-target reliability analysis pipeline used throughout the paper. Samples are first grouped by the estimated cleanliness score into ten equal-width bins over \([0,1]\), and we then summarize how pseudo-target behavior changes as samples move from low-clean to high-clean regimes.

\textbf{Panel (a): bin-wise pseudo-target mismatch.}
The horizontal axis is the midpoint of each cleanliness bin, and the vertical axis reports percentages. The blue curve (``Pseudo correct'') is the fraction of samples in that bin whose pseudo target matches the ground-truth class. The orange dashed curve (``Follow noisy label'') is computed only on observed-noisy samples and measures how often the pseudo target instead agrees with the noisy observed label. The shaded region marks the high-clean regime with bin midpoint at least \(0.65\). This panel is intended to show that once samples move into apparently clean regions, pseudo targets can still inherit the noisy label direction rather than becoming uniformly trustworthy replacements.

\textbf{Panel (b): aggregate failure across cleanliness regimes.}
The horizontal axis groups samples into three regimes: all samples, mid-clean bins \([0.35, 0.65)\), and high-clean bins \([0.65, 1.0]\). The vertical axis again reports percentages. The orange bars (``Follow noisy'') aggregate, over observed-noisy samples only, the rate at which the pseudo target agrees with the noisy observed label. The blue bars (``Pseudo correct'') aggregate the overall pseudo-target accuracy in that regime. The red bars (``High-conf wrong'') report the fraction of samples whose pseudo target is wrong despite satisfying the high-confidence criterion used by the reliability-analysis pipeline. The bracket in the high-clean regime highlights the gap between noisy-label following and actual pseudo-target correctness.

\textbf{Main takeaway.}
The central message of Figure~\ref{fig:intro_pseudo_target_unreliable} is not merely that noisy samples are hard; it is that high estimated sample cleanliness does not automatically imply high pseudo-target reliability. In the high-clean regime, the pseudo target follows the noisy observed label \(99.7\%\) of the time, while being correct on only \(74.7\%\) of samples and still making \(19.2\%\) high-confidence errors. This mismatch motivates the main paper's decision to treat observed-label reliability and pseudo-target reliability as related but distinct quantities.

\subsection{Details for Figure~\ref{fig:exp_pseudo_reliability}}
\label{app:fig3_details}

Figure~\ref{fig:exp_pseudo_reliability} reports the pseudo-target reliability diagnostics on CIFAR-100 with 50\% symmetric noise. The same trained baseline and TRACE-enhanced model are evaluated against the clean ground-truth labels, which are used only for analysis. Low-clean samples are those assigned low observed-label reliability, high-pconf samples are those whose pseudo targets pass the high-confidence criterion, and low-clean$+$noisy samples further restrict low-clean examples to cases whose observed label is actually corrupted.

\textbf{Panel (a): reliability by bin.}
Samples are grouped by clean-probability bins. The observed-clean curve measures how often the observed label is correct in each bin, while the pseudo-correct curve measures how often the selected pseudo target matches the clean class. The panel shows that pseudo-target correctness does not simply mirror observed-label cleanliness, motivating a separate pseudo-target reliability estimate.

\textbf{Panel (b): pseudo-target accuracy.}
This panel compares RoLR and TRACE across all samples, low-clean samples, high-pconf samples, and low-clean$+$noisy samples. The low-clean$+$noisy subset is the most diagnostic regime because the observed label is both unreliable and actually wrong, so refurbishment depends heavily on whether the pseudo branch is trustworthy.

\textbf{Panels (c) and (d): error quality.}
Panel (c) reports high-confidence pseudo-target errors, where lower values mean that fewer confident pseudo targets are wrong. Panel (d) reports pseudo-target calibration error. Together, these panels show whether the method only increases pseudo-target usage or also makes pseudo supervision safer.

\textbf{Main takeaway.}
Figure~\ref{fig:exp_pseudo_reliability} supports the main text's error-quality claim: TRACE improves pseudo-target accuracy most strongly in the low-clean$+$noisy regime, while also reducing high-confidence wrong pseudo targets and lowering calibration error. The gain is therefore not only a quantity increase in pseudo supervision, but an improvement in the reliability of the supervision that replaces corrupted labels.

\subsection{Details for Figure~\ref{fig:reliability_landscape}}
\label{app:fig5_details}

Figure~\ref{fig:reliability_landscape} visualizes pseudo-target accuracy over a two-dimensional reliability grid on CIFAR-100 with 50\% symmetric noise. The horizontal axis bins observed-label reliability into five intervals, and the vertical axis bins pseudo-target reliability into the same intervals. Each nonempty cell reports pseudo-target accuracy within that reliability pair; cells covering less than 0.01\% of samples are omitted to avoid over-interpreting extremely sparse regions.

\textbf{Panel (a): RoLR landscape.}
For RoLR, the observed axis is the original loss-based cleanliness score, while the pseudo axis reflects the pseudo-target confidence used for diagnostic binning. The landscape shows that low observed-label reliability alone does not guarantee reliable pseudo supervision: pseudo-target accuracy remains low in low-pseudo cells and becomes useful mainly when pseudo reliability is high.

\textbf{Panel (b): TRACE landscape.}
For TRACE, the observed axis uses the refined observed-label reliability score and the pseudo axis uses the separate pseudo-target reliability score. Compared with RoLR, TRACE improves the key low-observed / high-pseudo region, where pseudo supervision is most needed because the observed label is unreliable but a confident replacement is available.

\textbf{Main takeaway.}
The landscape explains why TRACE uses two gates rather than a single cleanliness coefficient. A sample should not move automatically from the observed-label branch to the pseudo branch just because observed-label reliability is low; it should receive strong pseudo supervision only in cells where pseudo-target reliability is also high.

\section{Plug-in Implementation and Computational Overhead}

\subsection{TRACE Pseudocode}

\begin{algorithm}[t]
  \caption{TRACE reliability interface for a mini-batch}
  \label{alg:trace_pseudocode}
  \small
  \begin{algorithmic}[1]
    \Require mini-batch \(\{x_i,\hat{\mathbf{y}}_i\}_{i=1}^{B}\), network predictions \(p_i^{(1)},p_i^{(2)}\), layer features \(\{h_l^{(1)}(x_i), h_l^{(2)}(x_i)\}_{l=1}^{L}\), loss-clean posterior \(c_i^{\mathrm{loss}}\)
    \Statex \textbf{Schedules/hyperparameters:} \(\beta_t,T,k,\gamma,\alpha,\lambda_{\mathrm{dis}},\rho,\varepsilon,w_{\min}\), pseudo-score start epoch
    \Ensure corrected target \(\tilde{\mathbf{y}}_i\), sample weight \(w_i\), weighted training loss \(\mathcal{L}\)
    \State \(q_i \gets \mathrm{Sharpen}_T\!\left(\frac{p_i^{(1)} + p_i^{(2)}}{2}\right)\) for all \(i\)
    \State \(\alpha_t \gets 1-\beta_t(1-\alpha)\)
    \For{\(m \in \{1,2\}\)}
      \For{\(l = 1,\dots,L\)}
        \State form cosine relation matrix \(A^{(l,m)}\) from normalized features \(\{h_l^{(m)}(x_i)\}_{i=1}^{B}\)
        \State keep self and top-\(k\) off-diagonal local affinities: \(R^{(l,m)} \gets \mathcal{S}_k(A^{(l,m)})\)
      \EndFor
      \For{each sample \(i\)}
        \State \(\delta_{i,m} \gets \sum_{l=1}^{L-1}\|R_i^{(l+1,m)} - R_i^{(l,m)}\|_2 / \sqrt{B}\)
        \State \(c_{i,m}^{\mathrm{str}} \gets \mathrm{Norm}\!\left(\exp(-\gamma \,\mathrm{Norm}(\delta_{i,m}))\right)\)
      \EndFor
    \EndFor
    \State \(\bar{c}_i^{\mathrm{str}} \gets \frac{1}{2}(c_{i,1}^{\mathrm{str}} + c_{i,2}^{\mathrm{str}})\) for all \(i\)
    \State \(g_i^{\mathrm{agr}} \gets 1\) if \(\arg\max p_i^{(1)} = \arg\max p_i^{(2)}\), else \(\lambda_{\mathrm{dis}}\)
    \State \(s_i^{\mathrm{obs}} \gets \mathrm{clip}\!\left((\alpha_t c_i^{\mathrm{loss}} + (1-\alpha_t)\bar{c}_i^{\mathrm{str}})\big((1-\beta_t)+\beta_t g_i^{\mathrm{agr}}\big), 0, 1\right)\)
    \State \(s_i^{\mathrm{pseudo}} \gets \left(\max_c q_{i,c}\right)^{\rho}\) after pseudo-score warm-up
    \State \(a_i \gets s_i^{\mathrm{obs}}, \quad b_i \gets (1-s_i^{\mathrm{obs}})s_i^{\mathrm{pseudo}}\)
    \State \(\tilde{\mathbf{y}}_i \gets \frac{a_i \hat{\mathbf{y}}_i + b_i q_i}{a_i + b_i + \varepsilon}\)
    \State \(w_i \gets \max(a_i + b_i, w_{\min})\)
    \State normalize \(\{w_i\}\) by the batch mean and optimize the base soft-target loss with the original prior regularizer
    \State \Return \(\tilde{\mathbf{y}}_i\), \(w_i\), and \(\mathcal{L}\)
  \end{algorithmic}
\end{algorithm}

\subsection{Plug-in Implementation Notes}

TRACE is implemented without changing the base method's pseudo-target generator or data pipeline. Given the base cleanliness score, it computes the shallow-anchored structural confidence, applies the dual-network agreement gate to refine observed-label reliability, filters pseudo targets with a separate confidence score, and combines the two branches through the unified supervision weight described in the main method section.

In the compact loss in \cref{eq:supervision_weighted_loss}, \(p_i\) denotes the prediction of the network currently being optimized. In two-network refurbishment pipelines, the same target correction and sample reweighting are applied symmetrically to the paired networks.

In the RoLR instantiation, the original GMM loss-clean posterior is reused as the loss-side input to observed-label reliability. The structural branch is activated after warm-up, computed from sparse local relations, and ramped in gradually so that early noisy features do not dominate the reliability estimate. The pseudo branch is activated with the same start epoch as the structural branch and uses only pseudo-target confidence, leaving pseudo-target construction itself unchanged. Thus, all ablation factors in the main paper modify only the reliability interface, not the underlying pseudo-label generator.

\subsection{Computational Overhead}
TRACE adds one batchwise relation computation for the structural confidence and a lightweight agreement check between the paired networks. The structural term is computed from the intermediate features already produced by the forward pass: for each analyzed layer, we normalize the \(B\) mini-batch features, form a \(B\times B\) cosine relation matrix, keep the self-affinity and top-\(k\) off-diagonal local affinities per sample, and compare sparse relation rows across depth. This requires no dataset-level neighbor graph, memory bank, or clean validation pass. Its temporary dense memory is \(O(LB^2)\) for \(L\) analyzed layers and batch size \(B\), and the retained sparse relations are \(O(LB(k+1))=O(LBk)\). With the default CIFAR-style setting \(B=64\) and \(k=50\), this overhead is small relative to storing backbone activations. The dual-network agreement gate reuses the predictions already available in two-network refurbishment pipelines, so it does not introduce an additional model or extra forward pass.

\section{Limitations and Future Extensions}

TRACE is designed for refurbishment-based pipelines with explicit pseudo-target construction, so its current evidence is strongest for image-classification benchmarks and baselines that expose such targets. The hardest real-noise setting, CIFAR-100N, also shows that reliability decoupling improves the difficult subset only modestly. Future extensions could test stronger backbones, additional refurbishment pipelines, and broader reliability signals beyond the shallow structural anchor used here.

Open-set label noise is a fundamentally different research setting from the closed-set noise considered in this work: some corrupted samples belong to classes outside the predefined label space, so the system must identify class-external samples rather than merely decide whether an in-distribution label or replacement target is reliable \cite{Wang2018d}. Extending TRACE to this setting therefore requires an explicit rejection mechanism and open-set sample identification, not a direct reuse of the current scores. We plan to pursue this direction by integrating open-set detection with the two-source reliability framework.


\end{document}